%% file: 0_acl_latex.tex
\pdfoutput=1

\documentclass[11pt]{article}

\usepackage[preprint]{acl}

\usepackage{times}
\usepackage{latexsym}

\usepackage[T1]{fontenc}

\usepackage[utf8]{inputenc}

\usepackage{microtype}

\usepackage{inconsolata}

\usepackage{graphicx}
\usepackage{booktabs}

\usepackage{dirtytalk} 
\usepackage{csquotes} 
\usepackage{amsmath}
\usepackage{enumitem}

\usepackage{multirow}
\usepackage{tabularx} 
\usepackage{array}
\usepackage{ltablex}

\usepackage{algorithm}
\usepackage{algorithmic}
\usepackage{amsmath}

\usepackage{tcolorbox}
\usepackage{float}
\usepackage{newfloat}
\usepackage[skip=2pt]{caption}

\usepackage{tcolorbox}
\tcbuselibrary{breakable}
\definecolor{lightblue0}{HTML}{e8eef4}
\definecolor{darkblue0}{HTML}{7a9cc0}

\usepackage{ltablex}
\keepXColumns

\DeclareFloatingEnvironment[
    fileext=lob,
    listname=List of Text Boxes,
    name=Prompt,                       
    placement=htbp,
]{textbox}

\definecolor{lightgreen}{HTML}{e8f4ea}
\definecolor{darkgreen}{HTML}{b8d8be}

\title{Discovering Conceptual Metaphors Across Topics and Media Types}

\author{
   Alexandria Leto\,\,
   Rohan Das\,\,
    \textbf{Juan Vásquez}\,\, \\
   \textbf{Abram Handler} \,\,
   \textbf{Maria Leonor Pacheco}\\
   University of Colorado Boulder \,\, \\
   \texttt{\{alexandria.leto, maria.pacheco\}@colorado.edu} \\
}

\begin{document}
\maketitle
\begin{abstract}
Conceptual metaphors guide our thinking and actions by allowing us to reason about more abstract experiences (e.g., paying taxes) in terms of more concrete or embodied experiences (e.g., carrying a physical load) \cite{lakoffMetaphorsWeLive2011}. It follows that different conceptual metaphors can result in different reasoning: framing \textit{paying taxes} as an \textit{investment in a community} rather than a \textit{physical load} leads to a very different outlook on taxation. Identifying the conceptual metaphors guiding a speaker or writer thus helps to reveal their framing of events. Though these metaphors can't be observed directly, groups of linguistic metaphors---metaphorical expressions as they appear in language---serve as evidence for them. Motivated by this, we present an unsupervised method that extracts linguistic metaphors from a corpus and uses a structured clustering approach to form groups corresponding to conceptual metaphors. Using this method, we point to key topical and framing differences in left- vs. right-leaning podcasts. For example, left-leaning podcasts tend to conceptualize media stories as a weapon, while right-leaning sources commonly discuss the economy as a system subject to vertical changes.


 
\end{abstract}

\input{1_intro}

\input{2_rw}

\input{3_framework}
\input{4_eval}
\input{5_cs}
\input{6_conclusion}


\bibliography{custom}

\appendix

\input{7_appendix}

\end{document}

%% file: 1_intro.tex
\section{Introduction}
Conceptual Metaphor Theory (CMT) suggests that metaphors are one of the primary structures guiding human thought. Elements of a more abstract target domain correspond to elements of a more concrete source domain, allowing people to reason about the target domain in terms of the source domain \cite{lakoffContemporaryTheoryMetaphor1992}. Because different source domains may characterize the same target domain in starkly different ways, understanding the conceptual metaphors guiding a writer's or speaker's thinking can provide an understanding of their framing of phenomena or events. 

\begin{figure}
    \centering
    \includegraphics[width=0.8\linewidth]{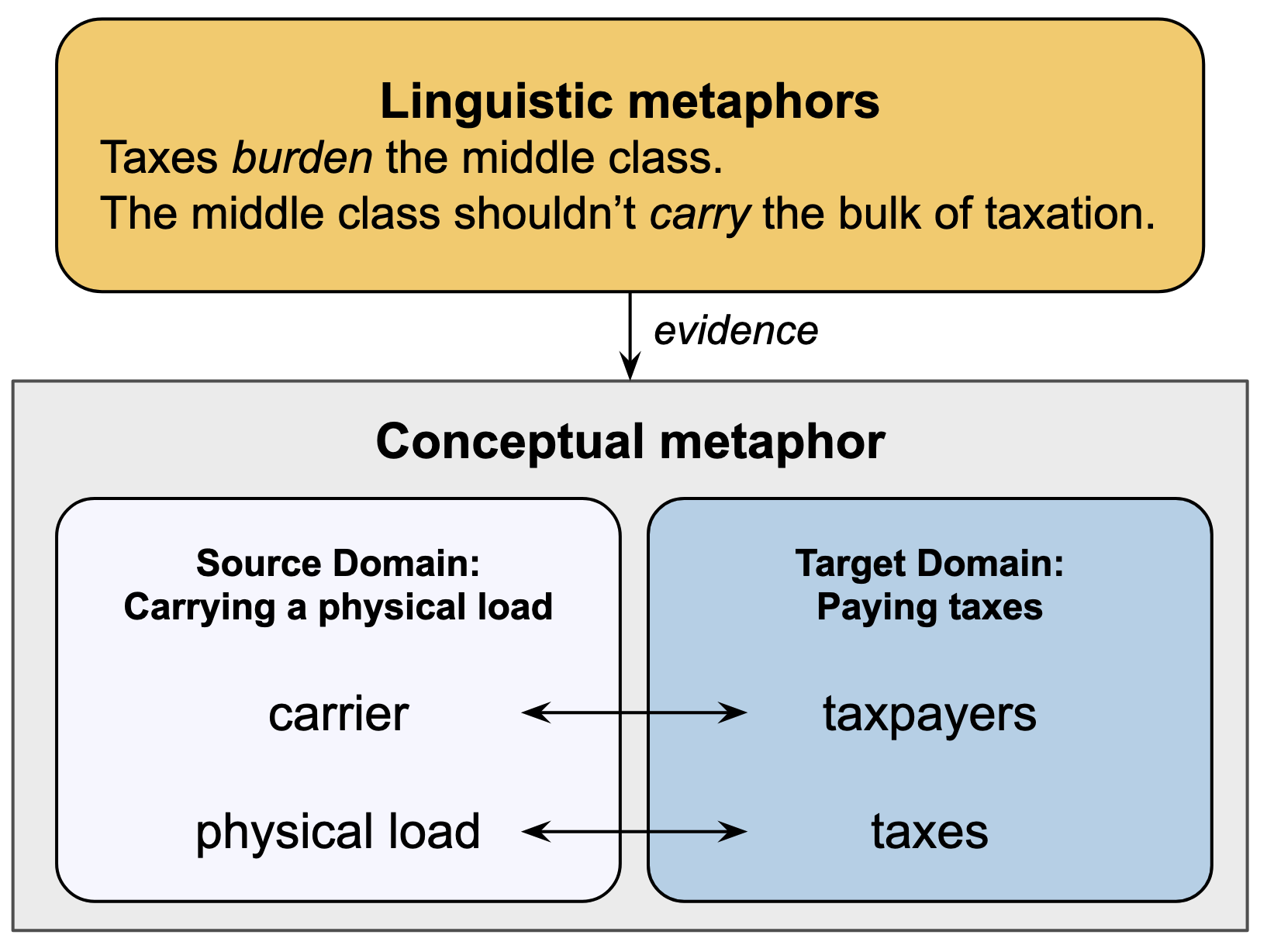}
    \caption{Example inspired by \citet{lakoffDontThinkElephant2004}.}
    \label{fig:taxation-example}
\end{figure}

For example, the \textit{paying taxes} target domain may be understood in terms of the \textit{carrying a physical load} source domain (see Fig. \ref{fig:taxation-example}). As a result, taxation will be treated as burdensome, obfuscating other potential outlooks, such as taxation as an investment in a community \cite{lakoffDontThinkElephant2004}. Because of this effect, identifying the conceptual metaphors guiding discourse can provide a rich dimension to media analysis. 


In spite of their usefulness for analysis, conceptual metaphors remain difficult to operationalize because they can't be observed directly. Instead, they are evidenced by their resulting linguistic realizations, known as \textit{linguistic metaphors}. For example, the phrases \say{taxes \textit{burden} the middle class} and \say{the middle class shouldn't \textit{carry} the bulk of taxation} both act as evidence for the \textit{paying taxes is carrying a physical load} conceptual metaphor. Recovering the conceptual metaphor from its realizations is what makes the analysis useful. Rather than simply understanding that a negative spin is being put on \textit{taxation}, we observe \textit{how} this bias is subtly communicated by following the logic of the \textit{carrying a physical load} source domain.

In this paper, we aim to identify and group linguistic metaphors to recover underlying conceptual metaphors in a discourse, thus providing a richer understanding of the framing within. 
To do this, we introduce an unsupervised method for inducing the conceptual metaphors in a corpus by identifying and grouping the linguistic metaphors within. Our pipeline first extracts all linguistic metaphors from a corpus. Then, we use a novel approach to generate rich interpretations for each. These linguistic metaphor interpretations include both a textual representation encoding the implied \say{framing effects} \cite{hoyleNaturalLanguageDecompositions2023, entmanFramingClarificationFractured1993} of the metaphor, as well as discrete properties of the source and target domains. 

Finally, we use a structured clustering approach. Grouping linguistic metaphors by surface similarity alone would separate realizations of the same conceptual metaphor whenever they appear in different contexts, and merge unrelated metaphors whenever they discuss the same issue. We therefore cluster over both components of the interpretation. Similarity is computed over the textual framing representations, while the discrete source and target properties are used as constraints on cluster assignment, so grouping is driven by framing effect and source--target structure rather than surface context.


To show that our induced groupings are indicative of media bias, we featurize the clusters and train a logistic regression classifier to predict political polarity based on cluster membership. Feature importance for this task reveals which clusters are characteristic of left-leaning versus right-leaning media, enabling a qualitative analysis of the conceptual systems of different groups of speakers or writers. We show that this method for inducing linguistic metaphor groups is feasible and accurate across topics and media types in a completely unsupervised setting. Upon confirming adequate performance, we then demonstrate our framework, showing that it facilitates a rich framing analysis of an unexplored corpus of podcast transcripts. 

All data and code will be released upon publication. Our contributions may be summarized as: (1) An unsupervised method which combines the generative abilities of LLMs with a structured clustering approach whose constraints are
grounded in CMT, which allows us recover conceptual metaphors guiding the framing present in a corpus. (2) A case study on an unexplored dataset of podcast transcripts which identifies conceptual metaphors used by right vs. left-leaning channels, providing preliminary evidence that left-leaning sources are more likely to characterize political violence metaphorically, while right-leaning sources are more likely to characterize economic issues.

%% file: 2_rw.tex
\section{Background}

\paragraph{Metaphor}
Metaphor has long been the subject of research in philosophy and linguistics, resulting in a variety of approaches and theories \cite{blackModelsMetaphors1962, hesseModelsAnalogiesScience1966, wilksPreferentialPatternseekingSemantics1975, barndenArtificialIntelligenceApproach2002, gentnerStructuremappingTheoreticalFramework1983}. 
CMT \cite{lakoffMetaphorsWeLive1980} has been particularly influential in computational metaphor studies \cite{martinComputationalModelMetaphor1990, fassMetMethodDiscriminating1991, barndenArtificialIntelligenceApproach2002, tianTheoryGuidedScaffolding2024}. Following these efforts, our work focuses on CMT, which originates with \citet{lakoffMetaphorsWeLive1980} and posits that the conceptual system governing a person's thoughts and actions is structured largely by metaphors. The theory does not treat linguistic metaphors in isolation, instead considering groups of linguistic metaphors as evidence for far-reaching patterns in a person's conceptual system; identifying these patterns can provide insights into the way someone thinks about, and thus frames, events or issues. This quality makes metaphors, as they are conceived in CMT, an important foothold for media analysis \cite{lakoffDontThinkElephant2004}. We expand on the theoretical grounding of our work in App. \ref{app:theo-grounding}.

\paragraph{Computational Metaphor Work}
Early computational approaches to metaphor detection and interpretation were rule-based, relying on extensive hand-coded knowledge \cite{martinComputationalModelMetaphor1990, fassMetMethodDiscriminating1991, barndenArtificialIntelligenceApproach2002, feldmanEmbodiedMeaningNeural2004, agerriMetaphorInferenceDomain2012}. These studies adhered closely to theories such as CMT, helping to advance an understanding of the structure of metaphor \cite{shutovaModelsMetaphorNLP2010}. 

To address the limited scope of hand-coded knowledge, a challenge in natural language processing more broadly, rule-based methods gave way to supervised learning, where a model is trained on a large hand-annotated corpus. This shift manifested as a focus on framing linguistic metaphor detection as a sequence classification task \cite{dodinhTokenLevelMetaphorDetection2016, swarnkarDiLSTMContrastDeep2018, suDeepMetReadingComprehension2020, chenGoFigureMultitask2020, gongIlliniMetIllinoisSystem2020}, as well as the development of large-scale annotated corpora to serve as ground truth training data \cite{mohlerIntroducingLCCMetaphor2016, krennmayrVUAmsterdamMetaphor2017}. As a result, metaphor interpretation efforts were de-emphasized \cite{pedinottiHowlingSuccessWorking2021}.

Recently, however, LLMs have been established as a viable option for generalizable metaphor detection without extensive hand-coded knowledge or annotated datasets \cite{tianTheoryGuidedScaffolding2024, puraivanMetaphorIdentificationInterpretation2024, linDualPerspectiveMetaphorDetection2025}. Focus has since shifted toward metaphor interpretation, a task LLMs are especially suited to given their generative capabilities~\cite{ichienLargeLanguageModel2024, tongMetaphorUnderstandingChallenge2024, sanchez-bayonaMetaphorLargeLanguage2025}.  

Given this renewed interest in interpretation, an intuitive direction is to design metaphor interpretation methods with theories such as CMT in mind, thus treating linguistic metaphors not as individual figures of speech, but instead as indicative of, and connected to, more schematic conceptual metaphors. This return to early work \cite{martinComputationalModelMetaphor1990, fassMetMethodDiscriminating1991, barndenArtificialIntelligenceApproach2002, feldmanEmbodiedMeaningNeural2004, agerriMetaphorInferenceDomain2012} could help develop a deeper understanding of the structural nature of metaphor. We argue that combining the capabilities of LLMs with constraint-based methods capable of explicitly modeling the structures proposed in theoretical work could be a path toward this goal. We see our method as an example of the feasibility and usefulness of this direction.

\paragraph{Metaphors and Computational Media Analysis}

Given that metaphor is considered a framing device \cite{burgersFigurativeFramingShaping2016}, 
metaphor has frequently been used to inform computational framing analysis. These works, often focused on dehumanizing metaphors in immigration discourse, typically measure the frequency of a pre-defined set of metaphorical source concepts \cite{mendelsohnFrameworkComputationalLinguistic2020, cardComputationalAnalysis1402022, mendelsohnWhenPeopleAre2025, otmakhovaNotAllANIMALs2026}. Though our work takes a broader view of metaphor, it is aligned with prior efforts in its shared goal of understanding the role of metaphor choice in media discourse.

%% file: 3_framework.tex
\section{Inducing Conceptual Metaphors}

\begin{figure}
    \centering
    \includegraphics[width=\linewidth]{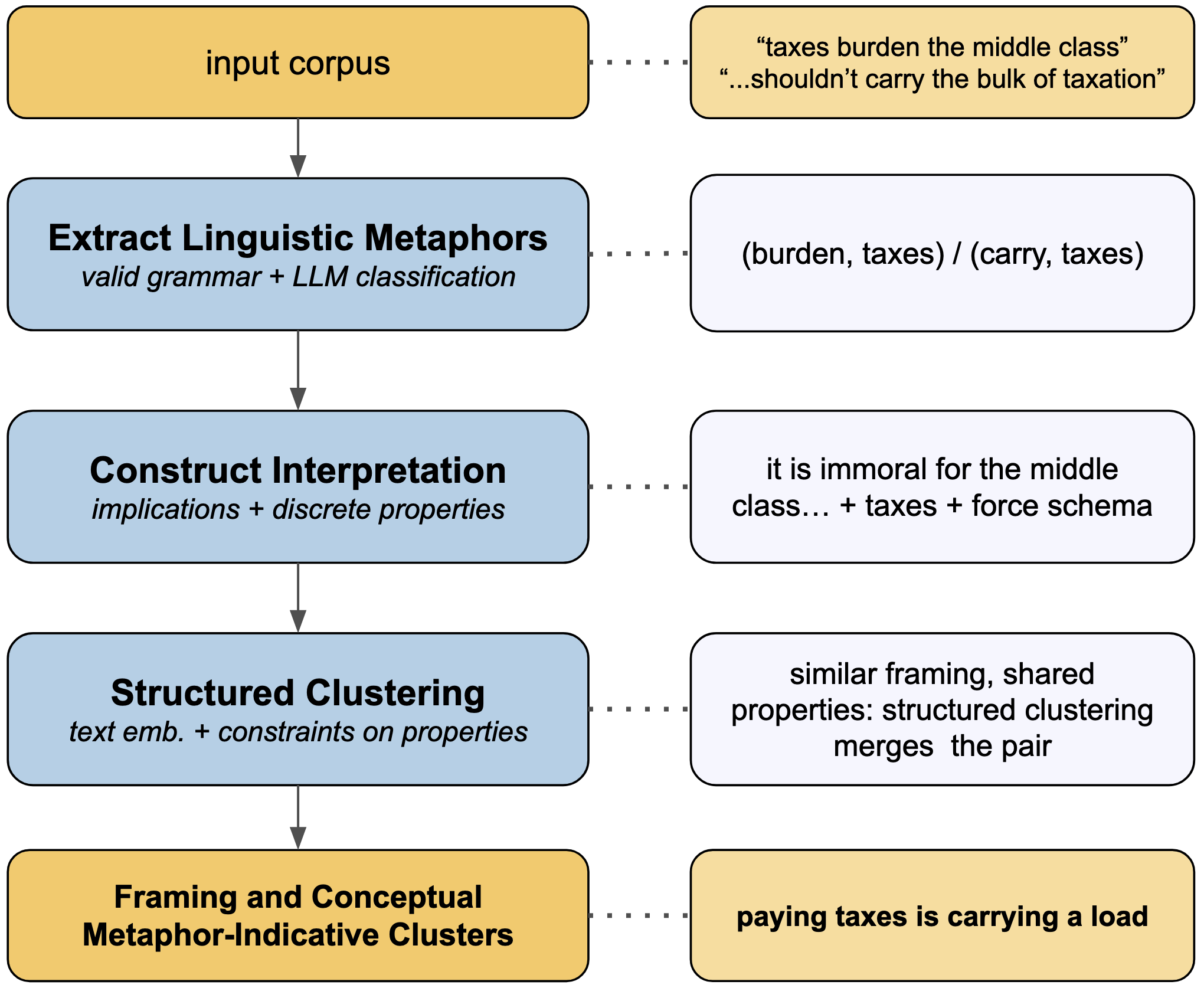}
    \caption{Pipeline for extracting linguistic metaphors and identifying framing-indicative groups.}
    \label{fig:pipeline}
\end{figure}

In CMT, linguistic metaphors are considered surface realizations of conceptual metaphors, or cross-domain mappings between a source and target domain. It follows that multiple distinct linguistic metaphors arise from the \textit{same} conceptual metaphor (Fig. \ref{fig:taxation-example}). Our framework is designed to gather evidence for pervasive conceptual metaphors driving the framing of an issue by constructing groups of linguistic metaphors likely arising from the same source--target domain mapping.  It first extracts linguistic metaphors from a corpus, then constructs an interpretation for each linguistic metaphor, and finally applies a structured clustering approach whose constraints discourage metaphors that differ on properties capturing the source and target sides of the mapping, so clusters are coherent in framing and domain structure. The full pipeline is summarized in Fig. \ref{fig:pipeline}.

\subsection{Extracting Linguistic Metaphors}
Before identifying high-level conceptual metaphors guiding the language in a corpus, we must first extract linguistic metaphors. We follow prior work \cite{birkeClusteringApproachNearly2006, gedigianCatchingMetaphors2006, shutovaMetaphorIdentificationUsing2010, shutovaUnsupervisedMetaphorIdentification2013, mohlerIntroducingLCCMetaphor2016} and consider linguistic metaphors to be within-sentence (source verb, target noun) pairs where the source verb is used metaphorically and indicates that a source domain is being used, while the target noun is an element of a target domain. For example, \textit{"targets burden the middle class"} $\rightarrow$ \textit{(burden, taxes)}

While other parts of speech can be metaphorical, 
we focus on verbs because they are more likely to be used metaphorically \cite{krennmayrVUAmsterdamMetaphor2017} and because LLMs perform consistently well on metaphorical verb identification \cite{puraivanMetaphorIdentificationInterpretation2024}. We focus on target nouns to align with prior work on dehumanizing metaphors, where the target word is typically a noun \cite{cardComputationalAnalysis1402022, mendelsohnWhenPeopleAre2025}.

Because CMT does not provide a systematic way to determine whether a word is metaphorical in a given context \cite{fassMetMethodDiscriminating1991, shutovaModelsMetaphorNLP2010}, we
consider a verb to be used metaphorically if its use transcends its original meaning without referring directly to a physical action \cite{puraivanMetaphorIdentificationInterpretation2024}. 


Building on prior work showing that linguistic metaphors occur in a limited set of grammatical patterns \cite{petruckMetaNetRepositoryIdentification2016}, we extract a set of candidate (source verb, target noun) pairs from each sentence in a corpus following the procedure in \citet{wangMetaphoricalFramingRefugees2024}. We identify all (noun, verb) pairs in each sentence, then use the spaCy python library \cite{honnibalSpaCyIndustrialstrengthNatural2020} to determine the shortest dependency path (SDP) between each pair. If the SDP matches one of the pre-defined construction patterns (shown in the Appendix, Tab. \ref{tab:con_patterns}), the corresponding target and source word pair is considered a \textit{metaphor candidate}. Then, to identify which target and source word candidates indeed constitute a linguistic metaphor, we prompt an LLM for a binary metaphor classification given the context sentence, target noun, and source verb. 


\subsection{Constructing Metaphor Interpretations}
\label{sec:text-rep}
Clustering the linguistic metaphor context sentences directly results in clusters overly focused on high-level topical information of the sentences, rather than their framing or source and target information (see App. \ref{app:sc-ablation}). Because we are interested in grouping linguistic metaphors with the same framing of events and source--target domains, we construct novel linguistic metaphor \textit{interpretations} to use for clustering. 

An interpretation has two components, corresponding to the two criteria we want clusters to share. The framing effects of a linguistic metaphor typically arise from what it \textit{implies}, rather than what is explicitly stated. To encourage clusters with similar framing, we first use an LLM to generate metaphor interpretations based on the metaphor's implied framing effects \citep{entmanFramingClarificationFractured1993}. This generated content is used for clustering. Separately, we elicit discrete LLM annotations of broad source and target properties, which are used to penalize cluster membership for linguistic metaphors that disagree on those properties. Section \ref{sec:struct-clust} describes how these are enforced.




\paragraph{Textual Interpretation}
We generate textual interpretations specifically focused on the way a metaphor \textit{frames} an event, which is often \textit{implied} by the metaphor rather than explicitly stated. Prior work shows that LLMs can effectively generate implicit context from an utterance \cite{hoyleNaturalLanguageDecompositions2023}. Inspired by this, we prompt in a few-shot setting for implications of each linguistic metaphor, guiding generation with questions corresponding to each of \citet{entmanFramingClarificationFractured1993}'s framing effects: (1) problem definition (e.g., taxes are burdensome), (2) causal interpretation (e.g., high tax rates cause taxes to be problematic), (3) moral evaluation (e.g., it is immoral for the middle class to carry the burden of taxes), and (4) treatment recommendation (e.g., \say{lightening the load} through reduced taxation).  

\paragraph{Discrete Property Annotations}
We obtain LLM annotations for discrete properties related to the target and source domains, respectively.

\noindent \textit{(1) Target Groups.}
Rather than annotating linguistic metaphors with target domains, we assign target \textit{group} labels identifying the class of entity the target belongs to. \textit{Taxpayers}, for instance, collects linguistic metaphors whose targets include the middle class, working families, and small business owners. This is a more straight-forward annotation task for an LLM than naming a conceptual domain, and it still encourages clusters sharing a target domain, since linguistic metaphors assigned to the same group are likely to draw on the same one.

To arrive at a canonical set of target groups for a corpus, we first prompt to classify each identified target noun as a person, place, thing, or organization. For each noun category, we cluster the contextualized noun embeddings. We sort the resulting clusters by proximity to the centroid, then prompt an LLM to identify the high-level noun groups within (e.g. immigrants, Democrats, the U.S.). We iteratively prompt the model to consolidate the list of groups by removing subsets and maintaining distinctions between groups only if they are important for discourse about a particular issue. This process yields a set of target groups, subdivided by person, place, thing, or organization as shown in App. Tab. \ref{tab:imm-tg}. Finally, we prompt to map each target word to a target group given its context sentence. If it does not fit cleanly into any of the domains, it is categorized as `other'. Additional details for this process may be found in App. \ref{app:target-domain-mapping}. 

\noindent \textit{(2) Image Schema Groups.}
Image schemata are recurring structures in our cognitive processes (e.g., \textit{containment}, \textit{path}) which act as basic building blocks of conceptual domains \cite{kovecsesDomainsSchemasFrames2020, oakleyImageSchemas2010, johnsonBodyMindBodily1990}. For example, the source domain \textit{carrying a physical load}, involves the \textit{counterforce} image schema, where two opposing forces are interacting. Like target groups, image schema groups are simpler for LLMs than source domains, and linguistic metaphors involving the same image schema are more likely to arise from the same source domain. Thus, we use image schema group labels to guide our clustering.

We prompt an LLM to identify the image schema group most associated with the literal context of each verb in our metaphorical set: spatial motion group (e.g., containment, path), force group (e.g., compulsion, counterforce), balance group (e.g., equilibrium), or other. The verb \say{burden}, for example, because it involves a \textit{counterforce} corresponds to the force group. These particular groups are chosen as they are well-developed by the introductory image schema work \cite{johnsonBodyMindBodily1990}. 

\subsection{Structured Clustering}
\label{sec:struct-clust}


We aim to go beyond simply clustering the textual interpretations of metaphors to produce clusters of metaphors with similar framings. Using the discrete properties above as clustering constraints, we aim to obtain more fine-grained groupings based on source--target domain mappings.

We adopt the constrained K-means clustering algorithm \cite{basuActiveSemiSupervisionPairwise2004} as implemented by \citealp{dasStructuredClusteringApproach2026}. We cluster the textual interpretations for the linguistic metaphors (see Section \ref{sec:text-rep}) under soft cannot-link constraints, which penalize placing a pair of linguistic metaphors in the same cluster when their target groups or image schema groups conflict. The reason we use only cannot-link constraints is that matching target and image schema groups is weak evidence that two linguistic metaphors realize the same conceptual metaphor, since many distinct mappings share coarse annotations, while conflicting groups are reliable evidence that they do not. The final output is a set of clusters of linguistic metaphors where each cluster is a plausible candidate set of evidence for a conceptual metaphor, as in Fig. \ref{fig:taxation-example}. Additional details are included in App. \ref{app:addl-sc}




%% file: 4_eval.tex
\section{Evaluation}
\label{sec:eval}
In this section we evaluate each component of our pipeline. We also summarize all potential failure points with examples in App. \ref{app:error-analysis}.

\noindent\textbf{Datasets.}
We use $\sim10,000$ samples from the LCC metaphor dataset, a collection of sentences spanning a variety of topics with source word, target word, and metaphor score annotations \cite{mohlerIntroducingLCCMetaphor2016} to benchmark our metaphor discovery component (Sec. \ref{subsec:eval-ling-mets}). This portion of the dataset is approximately balanced across positive and negative metaphor labels. We use datasets spanning three political discourse topics and two media types to demonstrate our framework's flexibility (Sec. \ref{subsec:rep-eval}, \ref{subsec:sc-eval}, and \ref{subsec:qual-eval}). The \textit{immigration} corpus is a dataset of $\sim1,600$ tweets about immigration issues \cite{mendelsohnWhenPeopleAre2025}. The \textit{gun control} and \textit{abortion} corpora are $1,000$ news articles each about the given topic~\cite{royWeaklySupervisedLearning2020}. All instances include political polarity (left/right) labels. Additional details for all datasets are include in App. \ref{app:addl-data-details}.

\noindent\textbf{Experimental Settings.} 
For all prompt-based methods we use Qwen3-8B \cite{yangQwen3TechnicalReport2025} with zero-shot and temperature $= 1$ unless otherwise stated. Prompts are included in App. \ref{app:prompts}. Results are reported on a single run. Detailed information about the annotation team is included in App. \ref{app:annotations}. 

\subsection{Quality of Linguistic Metaphors}
\label{subsec:eval-ling-mets}
We assess the quality of our extracted linguistic metaphors by benchmarking our metaphor detection framework on the LCC metaphor dataset.

\noindent\textbf{Metaphor Candidates.}
We sampled 100 candidate metaphor pairs (source verb, target noun) for evaluation by two annotators. For each, annotators assigned a binary score: 1 if the dependency between the two words was correctly extracted and the pair qualifies as a metaphor candidate (i.e., matches the syntactic patterns in App., Tab. \ref{tab:con_patterns}), and 0 otherwise. Averaged over annotators and samples, the extraction scored 0.9 out of 1. Inter-annotator agreement was excellent, with a K-$\alpha$ of 0.889. 

\noindent\textbf{Metaphor Detection.}
We evaluate our approach on binary metaphor classification. We compare against two methods: a random baseline that assigns uniform random classifications and a supervised RoBERTa classifier following \citet{wangMetaphoricalFramingRefugees2024}. RoBERTa ($p=81.74$, $r=80.90$, $f1=81.22$) and Qwen3-8b ($p=78.32$, $r=85.19$, $f1= 81.61$) both outperform the baseline ($p=58.16$, $r=51.02$, $f1=54.35$). Our prompting method outperforms the supervised baseline marginally while requiring no supervision. Because we are prioritizing linguistic metaphor coverage, we also prefer the higher recall obtained with Qwen.

To confirm the method's generalizability, we also evaluate it on a portion of the immigration dataset which had not been previously annotated with metaphor labels. This addresses concerns that the high performance observed could be due to the labels being \say{memorized} (i.e., that the LCC evaluation dataset was used for pre-training Qwen). We prompt the model to generate metaphor labels for the complete dataset, then sample 100 metaphor candidates (50 labeled literal and 50 labeled metaphorical by the model). An expert annotator blindly labeled each as metaphorical or not. Qwen3-8b attained an F1-score of 64.15, confirming sufficient generalizability. 



\label{subsec:rep-eval}
\subsection{Quality of Metaphor Interpretations}
To assess metaphor interpretations, we conduct an annotation study to determine the quality of the generated text implications, target group annotations, and source image schema group annotations. Because the structured clustering method relies on this content, it's important for it to be reliable. We sampled 100 interpretations from each of the three domains (immigration, abortion, and gun control) for the study. The codebook is included in App. \ref{app:codebooks}. 

Following the protocol in \citet{dasStructuredClusteringApproach2026}, two annotators evaluated each interpretation on a 5-point scale, assigning 0/1 scores for five criteria: plausibility of generated implications based on the original sentence, the relatedness of the implications to the linguistic metaphor, correctness of noun group assignment, correctness of target group assignment, and correctness of image schema assignment. The annotators adjudicated disagreements, arriving at a final set of 5-point scores, where 5 represents perfect interpretations and 0 represents fully incorrect ones. The resulting average scores  are 3.76/5 for the immigration corpus, 4.12/5 for gun control, and 4.23/5 for abortion.  To measure annotation quality, we report Gwet's AC1, as it's unaffected by the low variability in the individual criteria, most of which are 1. The Gwet's AC1 (calculated separately for the five criteria, then averaged) was 0.612 for immigration, 0.834 for gun control, and 0.873 for abortion.

Results indicate reliable interpretations for the gun control and abortion datasets, with immigration scoring somewhat lower. Inter-annotator agreement follows the same pattern, scoring high for gun control and abortion but moderate for the immigration corpus. The adjudication process revealed that this lower agreement was largely driven by a lack of calibration for one of the annotators. The high inter-annotator agreement for the other datasets supports that this was not an issue with the annotation protocol itself. Still, immigration quality scores are lower, even after adjudication. Closer examination shows that lower performance is primarily driven by errors on target group assignments, where we find that the model often assigned the \say{Anti-Immigration Stance} label to any linguistic metaphor with an anti-immigration framing, rather than focusing on the target noun itself. We interpret results on the immigration corpus with corresponding caution. A detailed breakdown of annotation statistics and results are included in App. \ref{app:addl-interp-results}.

\label{subsec:sc-eval}
\subsection{Quality of Clusters}

We assess whether our structured clustering produces coherent groups of linguistic metaphors and, in particular, whether the domain constraints improve on standard clustering. We first conduct an ablation study on the immigration corpus, evaluating clusters produced by vanilla K-means and by structured clustering with all possible combinations of constraints across the metaphor interpretation properties (noun, target, and image schema group). Our analysis reveals that applying cannot-link constraints to both the target group and image schema group yields the most consistent framing and source--target information within clusters (App. \ref{app:sc-ablation}). We adopt this constraint configuration for all subsequent experiments. We then use a systematic process (App. \ref{app:selecting-clusters}) to select the optimal number of clusters and constraint weights for each dataset, and evaluate our resulting clusters' purity with respect to target and source annotations, predictive power for political polarity, and performance on an intrusion test, 
across all three discourse topics. 

\noindent\textbf{Purity.}
We measure cluster purity with respect to the discrete target and source properties. For each cluster, we take the majority label as ground truth and average member agreement with it, over all clusters and label types.
Structured clustering improves cluster purity over K-Means in all cases (Tab. \ref{tab:sc-quant-perf}). This is expected, since our cannot-link constraints act directly on these properties. We report this to confirm that constraints behave as intended.


\begin{table*}[t]
\centering
\footnotesize
\renewcommand{\arraystretch}{0.95}
\setlength{\tabcolsep}{5pt}
\begin{tabular}{
  >{\raggedright\arraybackslash}p{0.08\linewidth}
  >{\raggedright\arraybackslash}p{0.25\linewidth}
  >{\raggedright\arraybackslash}p{0.58\linewidth}
}
\toprule
\textbf{Import. Ranking} & \textbf{Conceptual Metaphor} & \textbf{Linguistic Metaphors} \\
\midrule

1 &
\textbf{Economic gain is up/forward.} &
``Big Tech...IS \textit{putting} US workers last for greed \& power.'' \newline
``... [corporations] want to \textit{keep} money at the top.'' \newline
``They were so greedy they \textit{kept} America behind.'' \\
\midrule

8 &
\textbf{Immigration is a natural disaster.}  &
``...how about the US stops illegals from \textit{pouring} over our borders.'' \newline
``ILLEGALS have been \textit{flooding} US for so many years...'' \newline
``How can...CITIZENS \textit{stem} this tide of mass illegal immigration?'' \\
\midrule

20 &
\textbf{Immigrants are parasites.} &
``Millions of illegals will \textit{drain} our country including your bank account.'' \newline
``I am tired of \#illegals coming into America and \textit{sapping} the system dry'' \newline
``...keep illegals from crossing in...and \textit{sucking} the country dry'' \\

\bottomrule
\end{tabular}
\caption{Top-ranked metaphor clusters and framing for the immigration tweets corpus.}
\label{tab:immig-qual-analysis}
\end{table*}

\noindent\textbf{Predictive signal from clusters.}
One of our criteria for good clusters is cohesive framing of events. Because framing is highly indicative of political polarity, we assess the predictive power of a document's cluster membership for political polarity. We train a logistic regression classifier which takes a document-level cluster frequency vector as input. The vectors are of length $k$, the number of clusters, and each feature is initialized to 0. If the document contains a linguistic metaphor belonging to cluster $i$, feature $i$ is incremented by 1. Classifiers are evaluated using 5-fold cross-validation and their F1 scores for predicting political polarity are shown in Tab. \ref{tab:sc-quant-perf}. 
With only cluster membership as input and no textual features, the classifier's absolute scores are modest, yet predictive signal remains. This shows that which metaphor clusters a document draws on is itself informative about political polarity. Our structured clustering outperforms K-means, suggesting its constraints result in clusters that are better aligned with ideological framing.

\begin{table}[t]
\centering
    \resizebox{\columnwidth}{!}{%
        \begin{tabular}{l c l cc}
        \toprule
        \textbf{Topic} & \textbf{Clusters} & \textbf{Model} & \textbf{Polarity F1} & \textbf{Purity} \\
        \midrule
        
        \multirow{2}{*}{Immigration} & \multirow{2}{*}{225} & K-means & 51.92 & 20.85 \\
         & & Structured & \textbf{54.48} & \textbf{43.36} \\
        \midrule
        
        \multirow{2}{*}{Gun Control} & \multirow{2}{*}{175} & K-means & 48.28 & 17.53 \\
         & & Structured & \textbf{55.47} & \textbf{94.6} \\
        \midrule
        
        \multirow{2}{*}{Abortion} & \multirow{2}{*}{250} & K-means & 48.44 & 17.00 \\
         & & Structured & \textbf{52.88} & \textbf{95.47} \\
        \bottomrule
        \end{tabular}
    }
\caption{Performance comparison across topics for cluster purity and political polarity prediction.}
\label{tab:sc-quant-perf}
\end{table}

\noindent\textbf{Intrusion test.}
To assess the cohesiveness of the clusters, we conduct an intrusion test. We randomly sample 50 clusters, then for each, randomly select two linguistic metaphor interpretations from the top 50\% of interpretations nearest to the centroid. We randomly select an out-of-cluster interpretation to include with each ground-truth pair. Two annotators are instructed to identify the \say{intruder} or the out-of-cluster sample given only the linguistic metaphor interpretations. The logic behind this evaluation method is that if clusters are cohesive with respect to topic, framing, and source--target information, then identifying the \say{intruder} is intuitive. The accuracy, averaged over the two sets of annotations, was 82\% for immigration, 61\% for gun control, and 74\% for abortion. Inter-annotator agreement (K-$\alpha$) was 0.73 for immigration, 0.71 for gun control and 0.88 for abortion. The results are relatively stable for the immigration and abortion dataset. While performance on the gun control dataset is much lower, we find a significant difference in the performance of samples taken from the 25\% nearest the centroid (82\%) vs. the 25-50\% nearest the centroid (40\%), showing that linguistic metaphors nearer the centroid remain cohesive.

\subsection{Qualitative Analysis}
\label{subsec:qual-eval}

We evaluate the complete pipeline's viability for the downstream task: an analysis of clusters to gather linguistic metaphors serving as evidence for the conceptual metaphors. We focus on the immigration corpus, where we can determine whether we can recover well-documented metaphors in immigration discourse (e.g. immigrants as commodities, animals, parasites, etc.) \cite{santaanaAnimalWasTreated1999, obrienIndigestibleFoodConquering2003, cardComputationalAnalysis1402022, mendelsohnWhenPeopleAre2025} in addition to novel insights. We order the linguistic metaphor clusters by feature importance (App. \ref{app:addl-lr-details}) for predicting political polarity, as clusters with greater importance tend to be cleaner and more polarized. We follow the conventions of \citet{lakoffContemporaryTheoryMetaphor1992}, where conceptual metaphors are summarized as \textit{TARGET DOMAIN is SOURCE DOMAIN}. Our analysis is summarized in Tab. \ref{tab:immig-qual-analysis}.

We find linguistic metaphors evidencing the common conceptual metaphors \textit{immigration is a natural disaster} (cluster 8) and \textit{immigrants are parasites} (cluster 20). Recovering these known metaphors, particularly across different metaphorical verbs, gives us confidence that our pipeline is capable of unveiling insights beyond verb usage.


In line with this, cluster 1 shows how more general directional conceptual metaphors (\textit{good is up} and \textit{good is forward} \cite{lakoffMetaphorsWeLive1980}) are used to reason about economic gain. Here, U.S. corporations are framed as \say{greedy} entities who want to maintain the status quo by hoarding money among those already financially successful (at the top). Beyond these, our pipeline surfaces conceptual metaphors we did not anticipate, such as \textit{Gun ownership qualification is entering a building} in the gun control corpus and \textit{Opinions/stances are physical structures} in the abortion corpus (see App. \ref{app:addl-qual-analysis} for full analysis). By recovering well-known dehumanizing metaphors in addition to previously unremarked ones, we show that our pipeline effectively groups linguistic metaphors to uncover the conceptual metaphors structuring a discourse. 


        
         

        



%% file: 5_cs.tex
\section{Partisan Metaphors in Podcasts}
\label{sec:case-study}

In this section, we leverage our conceptual metaphor discovery pipeline to conduct an analysis of an unexplored dataset of podcast transcripts, effectively showing that our method gathers meaningful insights about partisan framing in novel datasets and diverse media types. 

Podcasts have become an important media format in the news media landscape, with around a third of U.S. adults, regardless of political affiliation, reporting that they receive news from podcasts \cite{shearerPodcastsNewsFact2025}. Podcast data provides a unique opportunity to analyze framing in a setting where political discourse is spoken and relatively informal. Motivated by this, we use our pipeline to understand how conceptual metaphors are used to guide framing of general U.S. political issues in this novel format.

\begin{figure}
    \centering
    \includegraphics[width=0.8\linewidth]{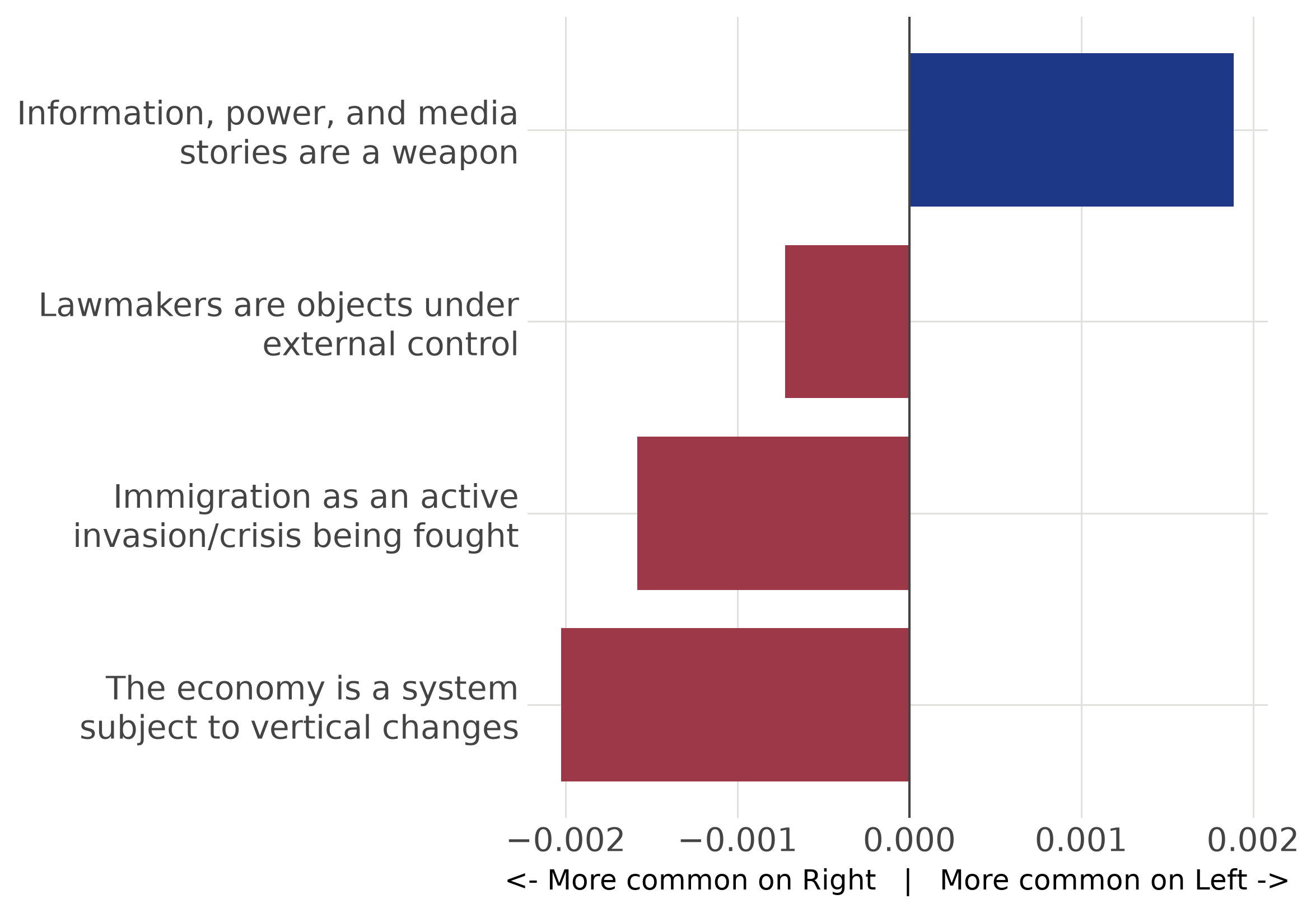}
    \caption{Conceptual metaphors in left vs. right sources.}
    \label{fig:lr-cm}
\end{figure}

\noindent\textbf{Dataset Construction.}
We identify five of the most popular U.S. political podcasts on the left and five on the right \cite{edisonresearchTopPodcastsConservative2024}, then sample and transcribe two shows from each year (2022-2026) where episode data is available, resulting in 94 episodes. 
We run our linguistic metaphor extraction on the episodes, then randomly sample 50 from each, for a total of 4,650 linguistic metaphors. 
We run the rest of our pipeline 
obtaining $k=225$ conceptual metaphor clusters for analysis. 
See App. \ref{app:addl-cs} for channel list and processing details.

\noindent\textbf{Analysis.}
We conduct a qualitative analysis of the conceptual metaphors most characteristic of left- and right-leaning podcasts (see Fig. \ref{fig:lr-cm}, full qualitative analysis in App. Tab. \ref{tab:cs-qual-analysis}). We find that among the most commonly invoked conceptual metaphors on the left is \textit{information/power as a weapon} through metaphorical usage of the verbs \textit{crush}, \textit{wield}, and \textit{seize}. On the right, the most prevalent include a conceptual metaphor well known in immigration discourse, in which immigration is discussed as an invasion.

\begin{figure}
    \centering
    \includegraphics[width=0.83\linewidth]{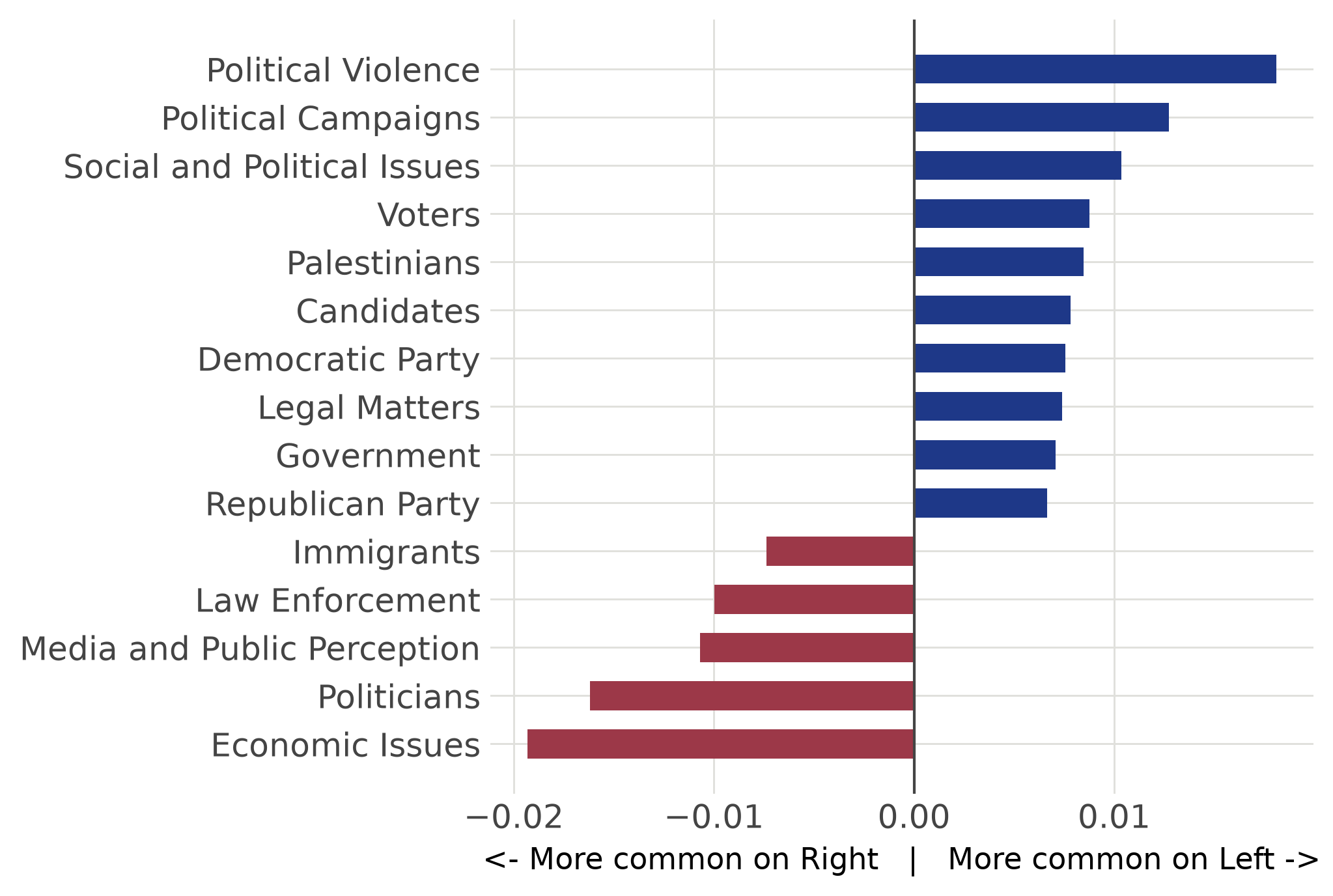}
    \caption{Most polarized target groups.}
    \label{fig:div-tg}
\end{figure}

Fig. \ref{fig:div-tg} shows the most divisive target groups in the dataset. We find that right-leaning sources are more likely to metaphorically characterize Economic issues and Politicians, while left-leaning sources are more likely to discuss Political Violence and Political Campaigns. A qualitative analysis of related linguistic metaphors shows that the left uses the Political Violence target group in a range of contexts including when discussing the January 6th insurrection, criticizing the right’s narrative of leftist political violence following Charlie Kirk’s assassination, criticizing ICE’s violence against protesters including the death of Alex Pretti and Renee Good, and discussing the conflict between Israel and Palestine. Right-leaning sources tend to use the Economic Issues target domain (often discussing it in terms of vertical shifts) to characterize the general economic status in the U.S. under different presidential administrations, the economic impact of immigration, and the U.S.'s reliance on Venezuelan oil.

%% file: 6_conclusion.tex
\section{Conclusion}
We present an unsupervised method for extracting linguistic metaphors from a corpus, then use a structured clustering approach to group linguistic metaphors to surface salient conceptual metaphors within, regardless of topic and media format. Across three domains, we show that structured constraints improve cluster quality over standard clustering, that cluster membership alone carries signal about political polarity, and that the resulting clusters are cohesive. 
Applying our framework to a dataset of U.S. news podcasts, we show that we can surface novel insights into how conceptual metaphors differ across the political spectrum.

\section{Limitations}
While we demonstrated that each component of our conceptual metaphor detection framework performs sufficiently well, our evaluation data includes only English-language data focused on U.S.-centric discourse and issues. Automatic methods necessarily result in some uncertainty.  We include an error analysis in Appendix Section \ref{app:error-analysis} to acknowledge failure points in our pipeline. Takeaways from the analysis must be considered with these possible errors in mind. 

We also acknowledge that, while the Whisper model used to transcribe the set of podcasts for the case study is widely used, we did not assess its accuracy in this work. The podcast dataset is also relatively small and covers a select set of podcast channels, so broad takeaways are limited.

\section{Ethical Considerations}
Because we rely on LLM-generated content for the linguistic metaphor interpretations, we acknowledge that our framework has the opportunity to amplify biases found in the LLM training data. To mitigate this, we use excerpts from the original data, rather than generated content, to guide all qualitative analysis.  We also acknowledge that, though a key goal of this work is to condemn dehumanizing language, we recognize the potential to reinforce biases by calling attention to them.

%% file: 7_appendix.tex
\section{Vocabulary from CMT}

Here we outline important terms that we use throughout this paper. 

\paragraph{Conceptual metaphor, cross-domain mapping, or metaphorical mapping.} \citet{lakoffContemporaryTheoryMetaphor1992} defines a conceptual metaphor using the \textit{LOVE IS A JOURNEY} conceptual metaphor as an example, stating: \say{a mapping (in the mathematical sense) from a source domain (in this case, journeys) to a target domain (in this case, love). The mapping is tightly structured. There are ontological correspondences, according to which entities in the domain of love (e.g., the lovers, their common goals, their difficulties, the love relationship, etc.) correspond systematically to entities in the domain of a journey} \cite{lakoffContemporaryTheoryMetaphor1992}.

\paragraph{Linguistic metaphor or metaphorical expression.} A linguistic metaphor is a metaphor as it appears in linguistic material such as a word, phrase, or sentence. It may be considered the \say{surface realization} of a conceptual metaphor \cite{lakoffContemporaryTheoryMetaphor1992}. 

\paragraph{Conceptual domain.} Source and target domains are conceptual domains. In this work and others \cite{lakoffContemporaryTheoryMetaphor1992}, they are considered the primary units that constitute a conceptual metaphor. Ontological correspondences occur between entities in \textit{conceptual domains} such as \say{journeys}, \say{love}, \say{time} and \say{money}. 

\paragraph{Image schema.} Image schemata are conceptual units considered more schematic, or abstract, than a conceptual domain \cite{kovecsesDomainsSchemasFrames2020}. They were first proposed in \citep{johnsonBodyMindBodily1990} and defined as follows: \say{...in order for us to have meaningful, connected experiences that we can comprehend and reason about, there must be a pattern and order to our actions, perceptions, and conceptions. An [image] schema is a recurrent pattern, shape, and regularity in, or of, these ongoing activities...I conceive of [image schemata] as structures for organizing our experience and comprehension}. Though conceptual domains are the primary unit we use to characterize conceptual metaphors, a conceptual domain is based on one or more image schemata. For example, the domain \say{journey} presupposes the image schema \say{motion}.

\section{Theoretical Grounding}
\label{app:theo-grounding}

In CMT, linguistic metaphors are considered surface realizations of conceptual metaphors, or cross-domain mappings between a source and target domain. It follows that multiple distinct linguistic metaphors arise from the \textit{same} conceptual metaphor (e.g., Fig. 1).

Conceptual metaphors guide the way a person reasons about, and therefore, frames issues \cite{lakoffMetaphorsWeLive2011}. Following \citet{entmanFramingClarificationFractured1993}, framing refers to the selection and emphasis of certain aspects of an issue to promote a particular interpretation. Because conceptual metaphors guide a writer's or speaker's reasoning \cite{lakoffMetaphorsWeLive2011}, they are a mechanism through which framing may be explained or characterized \cite{burgersFigurativeFramingShaping2016}. Linguistic metaphors can therefore be gathered to form evidence that a conceptual metaphor is shaping the framing of an issue. 

For example, if the linguistic metaphors \say{taxes burden the middle class} and \say{the middle class shouldn't carry the bulk of taxation} are extracted from a corpus, it is likely that the conceptual metaphor where \textit{carrying a physical load} is the source domain and \textit{paying taxes} is the target domain has informed the framing of taxation issues within.

Our framework is designed to gather evidence for pervasive conceptual metaphors driving the framing of an issue by constructing groups of linguistic metaphors likely arising from the same source---target domain mapping.  It first extracts linguistic metaphors from a corpus, second constructs a rich representation for each linguistic metaphor, and finally  utilizes a structured clustering approach to form cohesive groups.

Since its conception, a significant body of work has extended and clarified the original conception of CMT \cite{johnsonBodyMindBodily1990, kovecsesDomainsSchemasFrames2020, kovecsesBriefOutlineStandard2020, sullivanThreeLevelsFraming2023}. For example, \citet{kovecsesDomainsSchemasFrames2020} addresses the confusion between the various terms used to refer to the conceptual units involved in a conceptual metaphor. Where we have used \say{domains}, others have used image schemata and frames. The author collates the various units, arranging them into a hierarchy where image schemata are the most schematic, or abstract, then frames, and finally domains. In this work, we follow prior work \cite{lakoffContemporaryTheoryMetaphor1992} and refer to conceptual domains, or \say{a coherent area of conceptualization relative to which semantic units may be characterized} \cite{langackerFoundationsCognitiveGrammar1987} as the primary units constituting metaphors. Examples of conceptual domains include \textit{journeys}, \textit{time}, and \textit{war} \cite{lakoffContemporaryTheoryMetaphor1992}.

However, we also include image schemata annotations in our linguistic metaphor representations, which were first introduced by \cite{johnsonBodyMindBodily1990} and have been defined as \say{a condensed redescription of perceptual experience for the purpose of mapping spatial structure onto conceptual structure} \cite{oakleyImageSchemas2010}. They are more abstract and broad than domains, and based in bodily experience. Examples include \textit{container}, \textit{path}, and \textit{surface}. Image schemata help to make conceptual domains meaningful. For example, the \textit{journey} domain would be incomprehensible without the \textit{path} image schemata. Because of this effect, image schemata annotations are useful for guiding clustering without over-specifying the conceptual spaces involved in a given metaphor.



\section{Constructional Patterns for Candidate Metaphors}
The constructional patterns used to filter potential linguistic metaphors are shown in Fig. \ref{tab:con_patterns}.

\begin{table*}[h]
    \centering
    \small
    \begin{tabular}{@{}l l p{0.4\linewidth}@{}}
        \toprule
        \textbf{Pattern} & \textbf{Example} & \textbf{Full Example Phrase} \\
        \midrule
        S\_VERB $\xrightarrow{\textit{dobj}}$ T\_NOUN &
            attacked $\xrightarrow{\textit{dobj}}$ argument &
            He attacked the argument fiercely. \\[0.3em]
        T\_NOUN $\xrightarrow{\textit{nsubj}}$ S\_VERB &
            idea $\xrightarrow{\textit{nsubj}}$ collapsed &
            The idea collapsed under scrutiny. \\[0.3em]
        S\_VERB $\xrightarrow{\textit{agent}}$ ADP $\xrightarrow{\textit{pobj}}$ T\_NOUN &
            destroyed $\xrightarrow{\textit{agent}}$ by $\xrightarrow{\textit{pobj}}$ evidence &
            The theory was destroyed by the evidence. \\[0.3em]
        S\_VERB $\xrightarrow{\textit{amod}}$ T\_NOUN &
            crushing $\xrightarrow{\textit{amod}}$ argument &
            She made a crushing argument. \\[0.3em]
        T\_NOUN $\xrightarrow{\textit{nsubjpass}}$ S\_VERB &
            theory $\xrightarrow{\textit{nsubjpass}}$ shattered &
            The theory was shattered by new findings. \\
        \bottomrule
    \end{tabular}
    \caption{Construction patterns used to filter (noun, verb) pairs that can grammatically invoke a metaphor.}
    \label{tab:con_patterns}
\end{table*}

\section{Additional Target Domain Mapping Details}
\label{app:target-domain-mapping}
We prompt to classify the target noun appearing in each linguistic metaphor as a person, place, thing, or organization (Fig. \ref{prompt:target_noun_class_ann}). We extract the span embedding of the noun from its context sentence's RoBERTa \cite{liuRoBERTaRobustlyOptimized2019} embedding and use the K-means algorithm with cosine similarity to cluster the results. To identify the optimal $k$ value for this step, we select the $k$ value between 2 and 15 that yields the greatest silhouette score \cite{rousseeuwSilhouettesGraphicalAid1987}.

Within the resulting clusters, we sort the nouns and their context by proximity to the centroid. Next, we use a prompting method (Fig. \ref{prompt:create-target-groups}) to identify the high-level groups in each cluster before iteratively prompting to consolidate the list of groups, removing subsets and maintaining distinctions between groups only if they are important for discourse about a particular issue (Fig. \ref{prompt:clean-target-groups}). We repeat this final step until arriving at a set of $k$ target groups. The result is a set of target groups, subdivided by person, place, thing, or organization. Target groups for the immigration, gun control, and abortion datasets are included in App. \ref{app:tg-tabs}.

Finally, after establishing a set of target groups, we prompt an LLM to map each target word to a target group given its context sentence (Fig. \ref{prompt:target_group_ann}). If it does not fit cleanly into any of the domains, it is categorized as `other'. 

\section{Additional Structured Clustering Details}
\label{app:addl-sc}

To implement \cite{dasStructuredClusteringApproach2026}'s structured clustering approach, we obtain vector representations for the linguistic metaphors by embedding their textual representations (frame implications, see Section \ref{sec:text-rep}) using the SBERT all-MiniLM-L6-v2 model \cite{reimersSentenceBERTSentenceEmbeddings2019}. These are provided as inputs to the clustering algorithm. The constrained clustering objective adds a weighted penalty for instances in which a metaphor is assigned to a cluster that violates one or more of its associated cannot-link constraints. We also employ \citealp{dasStructuredClusteringApproach2026}'s constraint-aware centroid initialization algorithm. This algorithm preferentially selects initial centroids from instances subject to cannot-link constraints, thereby encoding constraint structure directly into the initialization step and yielding a more principled starting configuration for the subsequent clustering procedure. The structured clustering algorithm from \citet{dasStructuredClusteringApproach2026} is included in Alg. \ref{alg:constrained-kmeans}.

\begin{algorithm}[t]
\caption{Constrained k-means Clustering}
\label{alg:constrained-kmeans}
\begin{algorithmic}[1]
\STATE \textbf{Input:} $\mathcal{X} = \{x_i\}_{i=1}^n$ (SBERT embeddings of narrative chains), \\
       \hspace{3em} $\mathcal{C}$ (cannot-link constraints), \\
       \hspace{3em} $k$ (number of clusters), \\
       \hspace{3em} $w_{c}$ (constraint weight)
\STATE \textbf{Output:} Partitioning $\{\mathcal{X}_h\}_{h=1}^k$
\STATE \textbf{Method:}
\STATE Initialize $\{\mu_h^{(0)}\}_{h=1}^k$ via constraint-aware \\
       \hspace{1em} k-means++
\STATE $t \leftarrow 0$
\WHILE{not converged}
    \FOR{each $x_i \in \mathcal{X}$}
        \STATE Assign cluster:
        \STATE $h^* \leftarrow \arg\min_h \big[ \|x_i - \mu_h^{(t)}\|^2$ \\
        \hspace{3em} $+ w_{c} \sum_{(x_i,x_j) \in \mathcal{C}} \mathbf{1}[l_j^{(t)} = h] \big]$
        \STATE Assign $x_i$ to cluster $h^*$
    \ENDFOR
    \STATE Re-estimate centroids:
    \STATE $\mu_h^{(t+1)} \leftarrow \frac{1}{|\mathcal{X}_h^{(t+1)}|} \sum_{x_i \in \mathcal{X}_h^{(t+1)}} x_i$ \\
           \hspace{1em} for each cluster $h$
    \STATE $t \leftarrow t + 1$
\ENDWHILE
\end{algorithmic}
\end{algorithm}

\section{Additional Evaluation Dataset Details}
\label{app:addl-data-details}
Here we include additional information for each of the four datasets used for the evaluation in Sec. \ref{sec:eval}.

\paragraph{LCC}
We evaluate our metaphor detection component (Section \ref{subsec:eval-ling-mets}) using a portion of the English LCC dataset \cite{mohlerIntroducingLCCMetaphor2016}. In addition to source and target span annotations, each of the $\sim 78,000$ excerpts of the LCC dataset includes a metaphoricity score between 0 and 3 (inclusive). To convert metaphor scores to binary metaphor labels, we follow \citet{wangMetaphoricalFramingRefugees2024} and assign a negative label to those with a score of 0 and a positive score otherwise. The target concepts included in the dataset span a wide range of topics such as \say{guns,} \say{migration,} and \say{abortion} (see Tab. \ref{tab:lcc-target-concepts}). 
Because our method centers on identifying metaphorical (\texttt{source verb}, \texttt{target noun}) pairs, we focus on instances where the target and source spans include only single words. The source span must be a verb and the target span must be a noun. This results in a set of $\sim 10,000$ excerpts ($5,733$ positive and $4,285$ negative).

\paragraph{Immigration Tweets}
The original dataset of English tweets about immigration issues from \citet{mendelsohnWhenPeopleAre2025} contains two splits. We use the first, smaller split containing $\sim 1,600$ tweets for our evaluation. Each is annotated with political polarity and continuous metaphor scores between 0 and 1 indicating metaphoricity with respect to eight source domains: animal, commodity, parasite, pressure, vermin, war, water, and domain-agnostic. Because these metaphor scores are specific to these source domains, and our metaphor labels are source domain-agnostic, we do not use them in our evaluation.

The second split is larger, containing $\sim 35,000$ tweets. Because these tweets are not annotated with any metaphor-related labels, and thus would not appear in Qwen3-8b's training data, we use a portion of this split to analyze the generalizability of our prompting method to new documents.

\paragraph{Gun Control and Abortion News Articles}
The gun control and abortion datasets used for evaluation are both randomly sampled from the original datasets of $\sim6,000$ news articles each from \citet{royWeaklySupervisedLearning2020}. The datasets include article-level political polarity labels and the original gun control dataset spans 1984-2019, while the abortion dataset spans 1996-2019. 

\begin{table}[]
    \begin{center}
    \resizebox{0.6\columnwidth}{!}{%
        \begin{tabular}{lr}
            \toprule
            \textbf{LCC Target Domain} & \multicolumn{1}{l}{\textbf{Count}} \\
            \midrule
            guns & 2013 \\
            mental concepts & 1577 \\
            government & 1025 \\
            democracy & 673 \\
            elections & 613 \\
            bureaucracy & 603 \\
            poverty & 581 \\
            taxation & 498 \\
            wealth & 430 \\
            religion & 409 \\
            money & 313 \\
            disease & 231 \\
            migration & 152 \\
            intellectual property & 125 \\
            taxpayers & 104 \\
            taxes & 99 \\
            islamic & 86 \\
            terrorism & 85 \\
            gun debate groups & 79 \\
            gun rights & 78 \\
            politicians & 78 \\
            marriage & 45 \\
            drug trafficking & 40 \\
            welfare & 23 \\
            debt & 16 \\
            climate change & 15 \\
            abortion & 13 \\
            demographics & 11 \\
            control of guns & 3 \\
            \midrule
            \textbf{total} & 10018 \\
            \bottomrule
        \end{tabular}%
        }
    
    \end{center}
    \caption{Hand-annotated target concepts for all excerpts in the English LCC dataset used for evaluating the linguistic metaphor detection component of our framework. The dataset spans a wide variety of topics, making it an optimal resource for evaluating computational methods for general metaphor detection.}
    \label{tab:lcc-target-concepts}
    
\end{table}

\section{Annotations}
\label{app:annotations}
The annotation team used across annotation-based evaluation methods consisted of six annotators, three of whom are paper authors. The team included two experts holding PhDs in computer science, two computer science PhD students, one computer science Master's student and one linguistics Master's student. The team was balanced across gender identities, with 3 identifying as women and 3 as men. In all cases, the annotators were trained by the paper's first author and provided a codebook for the task, which is included in Appendix \ref{app:codebooks}.

\section{Quality of Extracted Linguistic Metaphors}
To evaluate binary metaphor classification performance on the LCC dataset (which includes four point metaphoricity scores), we follow \citet{wangMetaphoricalFramingRefugees2024} and assign negative labels to excerpts with a metaphoricity score of 0 and positive labels to those with a score of 1 or greater. Because we are interested in capturing even subtle metaphorical instances, this mapping aligns well with our goals.

The RoBERTa baseline follows \cite{wangMetaphoricalFramingRefugees2024}. The model is fine-tuned on the LCC dataset using source span embeddings as input to the classification layer. For this model, we use 5-fold cross-validation and report micro-averaged results.

Metaphor classification results are reported in Tab. \ref{tab:lcc_performance}.

\begin{table}[t]
    \begin{center}
        \resizebox{0.7\columnwidth}{!}{%
            \begin{tabular}{l|rrr}
                \toprule
                 & \textbf{precision} & \textbf{recall} & \textbf{f1} \\
                 \midrule
                \textbf{Random} & 58.16 & 51.02 & 54.35  \\             
                 \midrule
                \textbf{RoBERTa} & 81.74 & 80.90 & 81.22 \\
                \textbf{Qwen3-8b} & 78.32 & 85.19 & \textbf{81.61} \\
                 \bottomrule
            \end{tabular}%
        }
        \caption{Metaphor classification results LCC dataset.}
        \label{tab:lcc_performance}
    \end{center}
\end{table}

\section{Additional Interpretation Evaluation Results}
\label{app:addl-interp-results}

\paragraph{Inter-Annotator Stats}
A detailed breakdown of the annotation stats on the linguistic metaphor interpretation evaluation is shown for the immigration dataset in Table \ref{tab:i-agreement}, the gun control corpus in Table \ref{tab:gc-agreement}, and the abortion corpus in Table \ref{tab:a-agreement}. Annotation stats are relatively high, except for on the immigration dataset, where the adjudication process revealed that one of the annotators was poorly-calibrated, particularly on the source image schema group and target group portion of the task.

\begin{table}[]
    \begin{center}
    \resizebox{\columnwidth}{!}{%
    \begin{tabular}{lccc}
    \toprule
    \textbf{Annotation} & \textbf{K-$\alpha$} & \textbf{AC1} & \textbf{\%Agr.} \\
    \midrule
    Noun Group Score           & 0.162 & 0.912 & 92\%\\
    Target Group Score         & 0.532 & 0.550 & 77\% \\
    Source Img. Schema Group Score   & -0.015 & 0.286 & 58\%\\
    Interp. Plausibility Score  & 0.083 & 0.638 & 74\%\\
    Interp. Relevancy Score     & 0.292 & 0.667 & 73\%\\
    \bottomrule
    \end{tabular}
    }
        \end{center}
        \caption{Inter-annotator agreement scores across annotation categories for linguistic metaphor interpretations in the immigration corpus.}
\label{tab:i-agreement}
\end{table}

\begin{table}[]
    \begin{center}
    \resizebox{\columnwidth}{!}{%
    \begin{tabular}{lccc}
    \toprule
    \textbf{Annotation} & \textbf{K-$\alpha$} & \textbf{AC1} & \textbf{\%Agr.} \\
    \midrule
    Noun Group Score           & 0.884 & 0.989 & 99\% \\
    Target Group Score          & 0.697 & 0.802 & 88\% \\
    Source Img. Schema Group Score   & 0.472 & 0.845 & 88\% \\
    Interp. Plausibility Score & 0.454 & 0.812 & 86\% \\
    Interp. Relevancy Score    & 0.750 & 0.919 & 93\% \\
    \bottomrule
    \end{tabular}
    }
    \end{center}
        \caption{Inter-annotator agreement scores across annotation categories for linguistic metaphor interpretations in the gun control corpus.}
    \label{tab:gc-agreement}
\end{table}

\begin{table}[]
    \begin{center}
    \resizebox{\columnwidth}{!}{%
    \begin{tabular}{lccc}
    \toprule
    \textbf{Annotation} & \textbf{K-$\alpha$} & \textbf{AC1} & \textbf{\%Agr.} \\
    \midrule
    Noun Group Score                  & 0.388 & 0.968 & 97\% \\
    Target Group Score                & 0.690 & 0.805 & 88\% \\
    Source Img. Schema Group Score   & 0.416 & 0.648 & 78\% \\
    Interp. Plausibility Score & 0.951 & 0.987 & 99\% \\
    Interp. Relevancy Score    & 0.540 & 0.761 & 81\% \\
    \bottomrule
    \end{tabular}
    }
    \end{center}
        \caption{Inter-annotator agreement scores across annotation categories for linguistic metaphor interpretations in the abortion corpus.}
    \label{tab:a-agreement}
\end{table}

\paragraph{Detailed Evaluation}
Full per-interpretation criteria results for the immigration dataset are in Tab. \ref{tab:i-interp_scores}, the gun control dataset in Tab. \ref{tab:gc-interp_scores}, and the abortion dataset in Tab. \ref{tab:a-agreement}. We find that, like annotation, performance is stable except for on the immigration dataset. Lower performance is primarily driven by errors on target group assignments, where closer analysis revealed that the model was often assigning the \say{Anti-Immigration Stance} label to any linguistic metaphor with an anti-immigration framing, rather than focusing on the target noun itself.

\begin{table}[h]
\centering
\begin{tabular}{lc}
\hline
\textbf{Annotation} & \textbf{Score} \\
\hline
Noun Group Score & 0.95/1 \\
Target Group Score & 0.5/1 \\
Source Img. Schema Group Score & 0.8/1 \\
Interp. Plausibility Score & 0.75/1 \\
Interp. Relevancy Score & 0.76/1 \\
\hline
\end{tabular}
\caption{Scores for metaphor interpretations in the immigration dataset.}
\label{tab:i-interp_scores}
\end{table}

\begin{table}[h]
\centering
\begin{tabular}{lc}
\hline
\textbf{Annotation} & \textbf{Score} \\
\hline
Noun Group Score & 0.97/1 \\
Target Group Score & 0.76/1 \\
Source Img. Schema Group Score & 0.75/1 \\
Interp. Plausibility Score & 0.88/1 \\
Interp. Relevancy Score & 0.76/1 \\
\hline
\end{tabular}
\caption{Scores for metaphor interpretations in the gun control dataset.}
\label{tab:gc-interp_scores}
\end{table}

\begin{table}[h]
\centering
\begin{tabular}{lc}
\hline
\textbf{Annotation} & \textbf{Score} \\
\hline
Noun Group Score & 0.94/1 \\
Target Group Score & 0.72/1 \\
Source Img. Schema Group Score & 0.87/1 \\
Interp. Plausibility Score & 0.86/1 \\
Interp. Relevancy Score & 0.84/1 \\
\hline
\end{tabular}
\caption{Scores for metaphor interpretations in the abortion dataset.}
\label{tab:a-interp_scores}
\end{table}

\section{Structured Clustering Ablation Study}
\label{app:sc-ablation}

Here we conduct an ablation study on the structured clustering setup to determine the most effective method for constructing linguistic metaphor clusters. We use the immigration discourse dataset from \cite{mendelsohnWhenPeopleAre2025}. Because quantitative metrics for measuring clustering are relatively stable across setups (see Table \ref{tab:ablation-quant-results}), we conduct a qualitative analysis focused on the important clusters for predicting the political polarity of a tweet author. We evaluate them for their consistency of source and target information as well as framing of issues. This analysis is summarized in Table \ref{tab:ablation-qual-summary}.

We find that the clusters constructed using cannot-link constraints on image schema group and target group result in the cleanest clusters with respect to topic and framing. While the metaphorical consistency with respect to source and target characteristics may be slightly weaker than other setups, this enables groupings between various source and target concepts which may belong to the same conceptual domain.

\begin{table*}[t]
    \centering
    \begin{tabular}{l l c}
    \toprule
    \textbf{Clustering Method} & \textbf{Constraints} & \textbf{LR F1} \\
    \midrule
    kmeans (linguistic met. only) & --- & 56.27 \\
    kmeans (textual met. interpretation) & --- & 57.49 \\
    \midrule
    
    \multirow{4}{*}{pckmeans (met. interpretation)} & image schema group & 58.77 \\
    & target group & 59.04 \\
    & image schema group, noun classification & 58.15 \\
    & image schema group, target group & 54.48 \\
    \bottomrule
    \end{tabular}
    \caption{Logistic regression F1 for predicting political polarity in the immigration dataset, by clustering setup.}
    \label{tab:ablation-quant-results}
\end{table*}


\section{Selecting Clusters for Evaluation}
\label{app:selecting-clusters}
To select the optimal clustering hyperparameters ($k$, the number of clusters and $w_c$, the constraint weights) to use for evaluation, we roughly follow the process from \citet{dasStructuredClusteringApproach2026}. We evaluate polarity prediction performance and purity across a range of cluster sizes $K\in\{100, 125, 150, 175, 200, 225, 250, 275, 300\}$ and compare to vanilla k-means as a baseline. We select hyperparameters that optimize performance for both metrics. We find that $k=225$ and $w_c=0.05$ is best for immigration, $k=175$ and $w_c=0.1$ is best for gun control, and $k=250$ and $w_c=0.1$ is best for abortion.

\section{Additional Logistic Regression Details}
\label{app:addl-lr-details}
For the logistic regression classifier used for predicting document-level political polarity based on cluster membership, we split the documents into train/test sets consisting of 80\% and 20\% of the documents, respectively. We use scikit-learn's \cite{pedregosaScikitlearnMachineLearning2011} implementation of logistic regression with the `lbfgs' solver and conditional log-likelihood with $L_2$ regularization. We do not include significance testing because we evaluate with repeated stratified 5-fold, repeating 5 times and reporting mean performance. 

\section{Additional Qualitative Analysis}
\label{app:addl-qual-analysis}
Here we expand on our qualitative analysis of clusters resulting from our pipeline with the goal of demonstrating its usefulness for gathering linguistic metaphors to be used evidence for conceptual metaphors guiding framing in a corpus.

\paragraph{Immigration}
In addition to the \textit{immigration is a natural disaster} and \textit{immigrants are parasites} metaphors, we also recover a relatively clean \textit{immigrants are commodities/cargo} cluster (ranked 56). This includes the phrases: 

\begin{figure}[H]
\centering
\small
\begin{displayquote}
\say{Migrant Paid To ‘\textit{Rent}’ A Child To Help Him Cross The Border}\\[8pt]
\say{Since \textit{stealing} children from their parents FAILED...}\\[8pt]
\say{More worried about the immigrant children we kidnapped from their parents that are being \textit{sold} by contractors into adoption}
\end{displayquote}
\end{figure}

Though these phrases contain dehumanizing language, the framing with respect to immigration is bipartisan, which explains the relatively lower importance ranking. This is highly related to other work showing that dehumanizing metaphors are used by those with both pro- and anti-immigration stances \cite{mendelsohnWhenPeopleAre2025}.

We also find messier clusters with respect to conceptual metaphor consistency for well-documented metaphor types because some of the corresponding these linguistic metaphors are driven by parts of speech \textit{other than verbs}. For example, cluster 75 includes the phrases:

\begin{figure}[H]
\centering
\small
\begin{displayquote}
\say{'I'm thinking out loud but that would also heighten the \textit{hunt} for immigrants...}\\[8pt]
\say{these are \textit{'animals'}??wtf THESE ARE HUMAN BEINGS WHO WANT A BETTER LIFE...}\\[8pt]
\say{We say no to putting immigrant families in \textit{cages}...}\\
\end{displayquote}
\end{figure}

Each corresponds to the dehumanizing \textit{immigrants are animals} metaphor.

There are also clusters where a conceptual metaphor is not consistent, but framing is, leading to interesting and specific takeaways. For example, cluster 43 is largely a discussion of immigration issues in California where California's lawmakers are framed as \say{taking advantage} of undocumented immigrants for votes, thus prioritizing immigrants over California natives.  

\paragraph{Gun Control}
The qualitative analysis on the Gun Control corpus is summarized in Tab. \ref{tab:gc-qual-analysis}. We recover three robust conceptual metaphors shaping gun control discourse. The first is \textit{Gun ownership qualification is entering a building} (importance rank 1). The examples from this cluster specifically discuss background checks as fortifying the \say{building} by using verbs such as \say{strengthen}, \say{expand} and \say{close.} Improved background checks are proposed or criticized as a solution to widespread gun violence. The majority of examples are pulled from left-leaning sources. Another cluster (2), includes several linguistic metaphors evidencing a \textit{political disagreements are a physical struggle/war}. The phrases are bipartisan discussions of tension between the Democratic and Republican parties. In cluster 4, we see evidence of a \textit{gun ownership rights are an object on a scale}. These examples frame the process of developing gun regulations as a balancing act where public safety and 2nd-amendment rights are on either side of a scale. 

\paragraph{Abortion}
The qualitative analysis on the abortion corpus is summarized in Tab. \ref{tab:abor-qual-analysis}. In the most important cluster for predicting political polarity, we gather evidence for the \textit{opinions/stances are physical structures} from left-leaning stances. These examples frame anti-abortion stances as objects that are \say{supported}, \say{formed}, and \say{maintained} by those opposing abortion. In cluster 3, we identify the \textit{science and pro-choice stances are competing destructive forces}. The examples frame modern science as force that \say{demolishes} pro-abortion stances. Cluster 5 evidences the \say{media attention or stories are physical objects} to be \textit{caught}, \textit{revealed}, or \textit{buried}. 

\section{Error Analysis}
\label{app:error-analysis}

Examples for each of the possible framework errors (outlined below) may be found in Tab. \ref{tab:error-examples}.

\paragraph{Incorrect metaphor candidate extraction}
Due to parsing issues, the (target word, source word) pairs are not always accurately extracted: the shortest dependency path between the two words is sometime incorrect, resulting in a candidate pair that does not match any of the constructional patterns shown in Tab. \ref{tab:con_patterns}. The result is generally that the target word is in no way being characterized by the metaphorical source verb. 

\paragraph{Incorrect metaphor classification}
Though the performance of the metaphor classification component is generally good, it sometimes over-classifies. This tendency is consistent with other prompt-based methods used for metaphor classification. As a result, generated implications tend to focus broadly on the context sentence rather than the metaphor itself. This adds some noise with respect to source---target information in the downstream clusters.

\paragraph{Incorrect target group classification}
Relatively frequently, the LLM fails to accurately assign a target group label specifically to a noun, instead resulting in a broad topical label for the sentence. For example, while the target noun may refer to a politician, the model assigns an 'Immigrant' label simply because immigration is discussed in the sentence. The model also sometimes focuses on the incorrect noun in the sentence.

\paragraph{Incorrect image schema group classification}
At times, image schema group assignments are sorted into the `Other' category rather despite `Force' or `Spatial Motion' being the clearly correct choice. 

\paragraph{Incorrect frame implication generation}
The model can fail to generate implications aligned with the context sentence. When there is not enough context to determine the true implications of a sentence, the model can hallucinate. In fewer cases, the model generates implications that are not aligned with one another. For example, one implication may carry one spin while another implication carries the opposite.

\paragraph{Irrelevant frame implication generation}
The generated metaphorical implications can fail to specifically address the metaphorical verb, resulting in broad implications of the context sentence. Similar to metaphor misclassification, this can add noise to the resulting clusters.


\section{Additional Case Study Details}
\label{app:addl-cs}

\begin{figure}
    \centering
    \includegraphics[width=\linewidth]{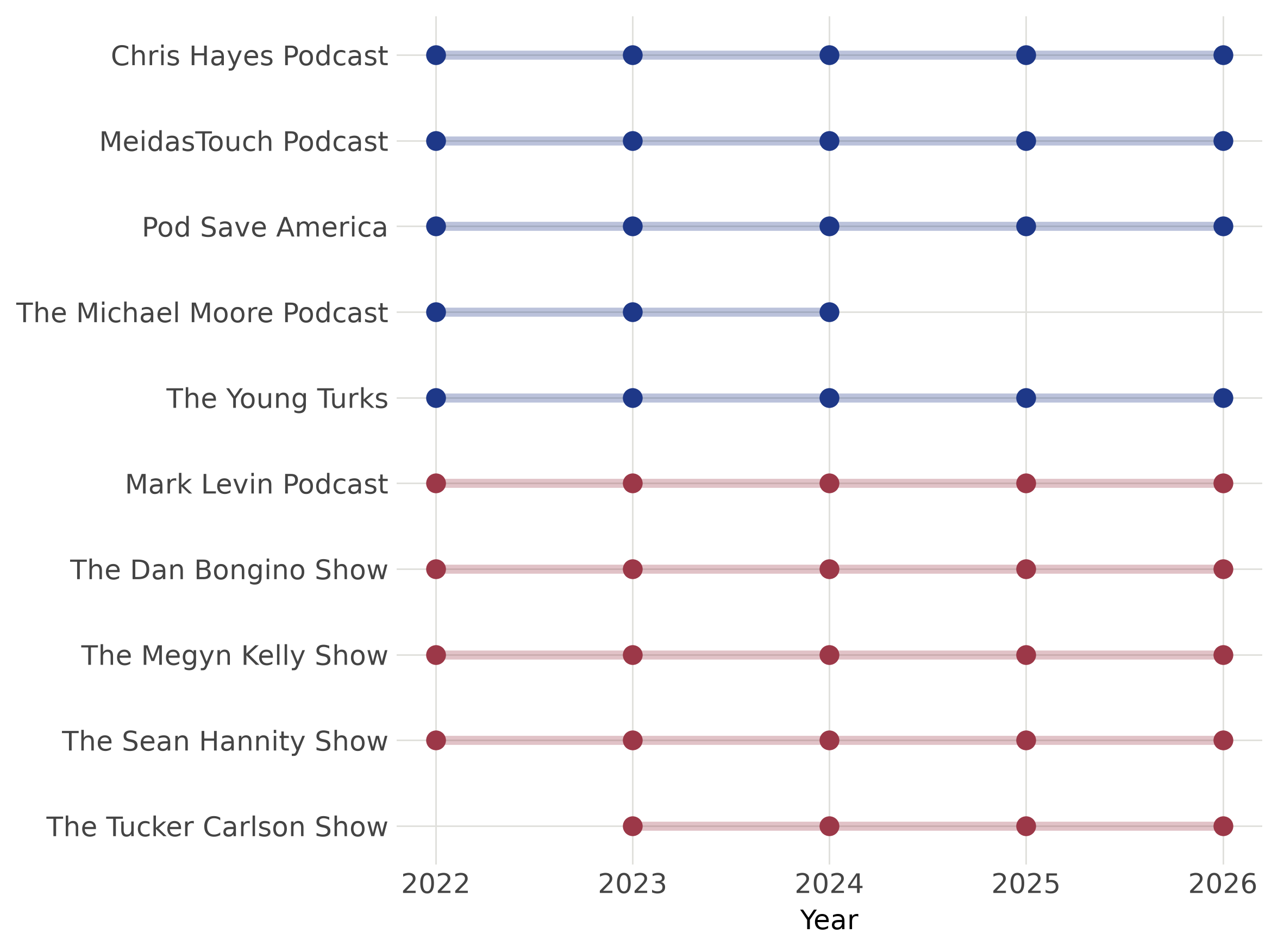}
    \caption{Channel coverage over analysis years.}
    \label{fig:cov-channel-year}
\end{figure}

\begin{figure}
    \centering
    \includegraphics[width=\linewidth]{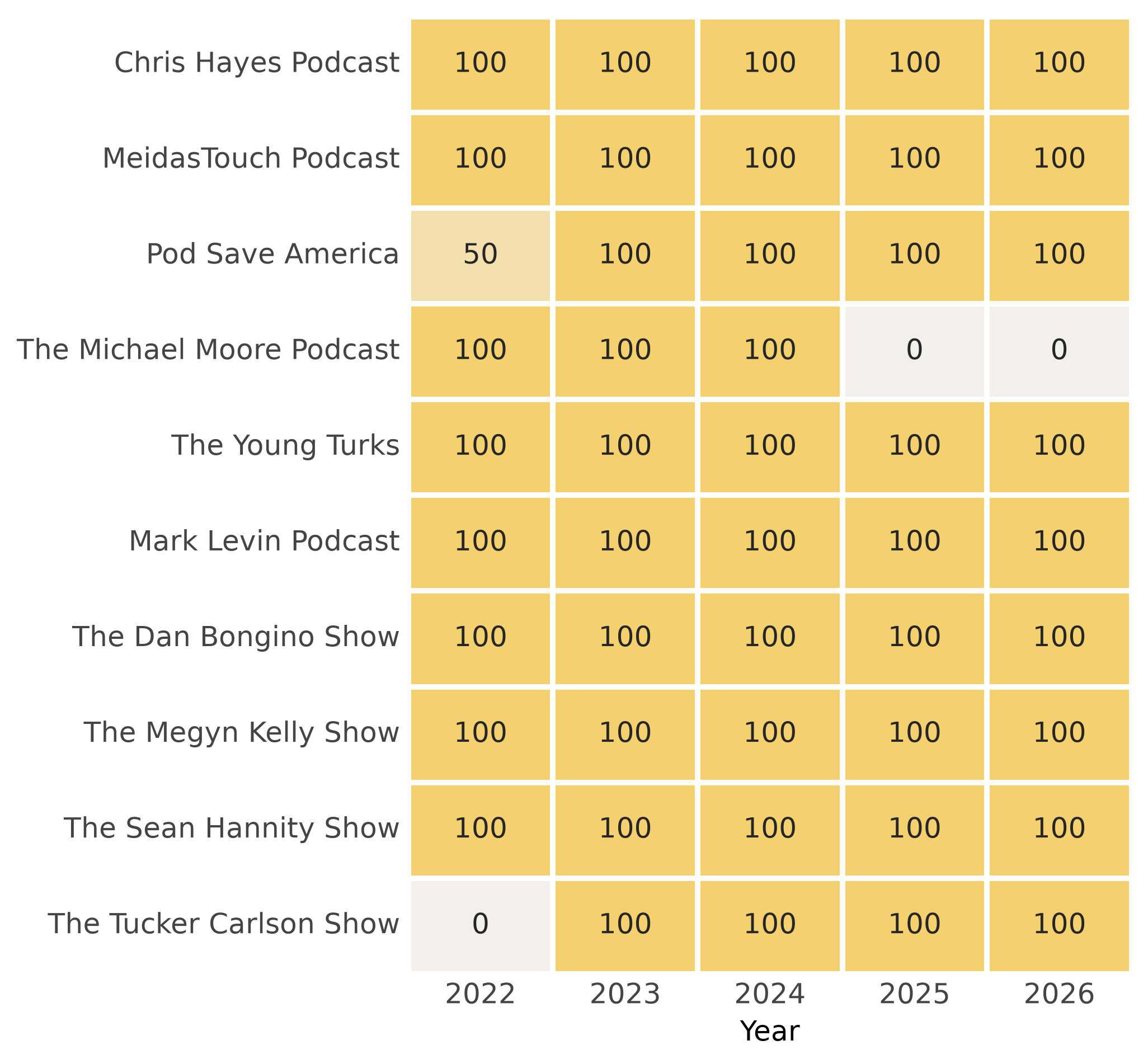}
    \caption{Number of linguistic metaphors sampled for each channel-year.}
    \label{fig:mets-channel-year}
\end{figure}

\begin{figure}
    \centering
    \includegraphics[width=1\linewidth]{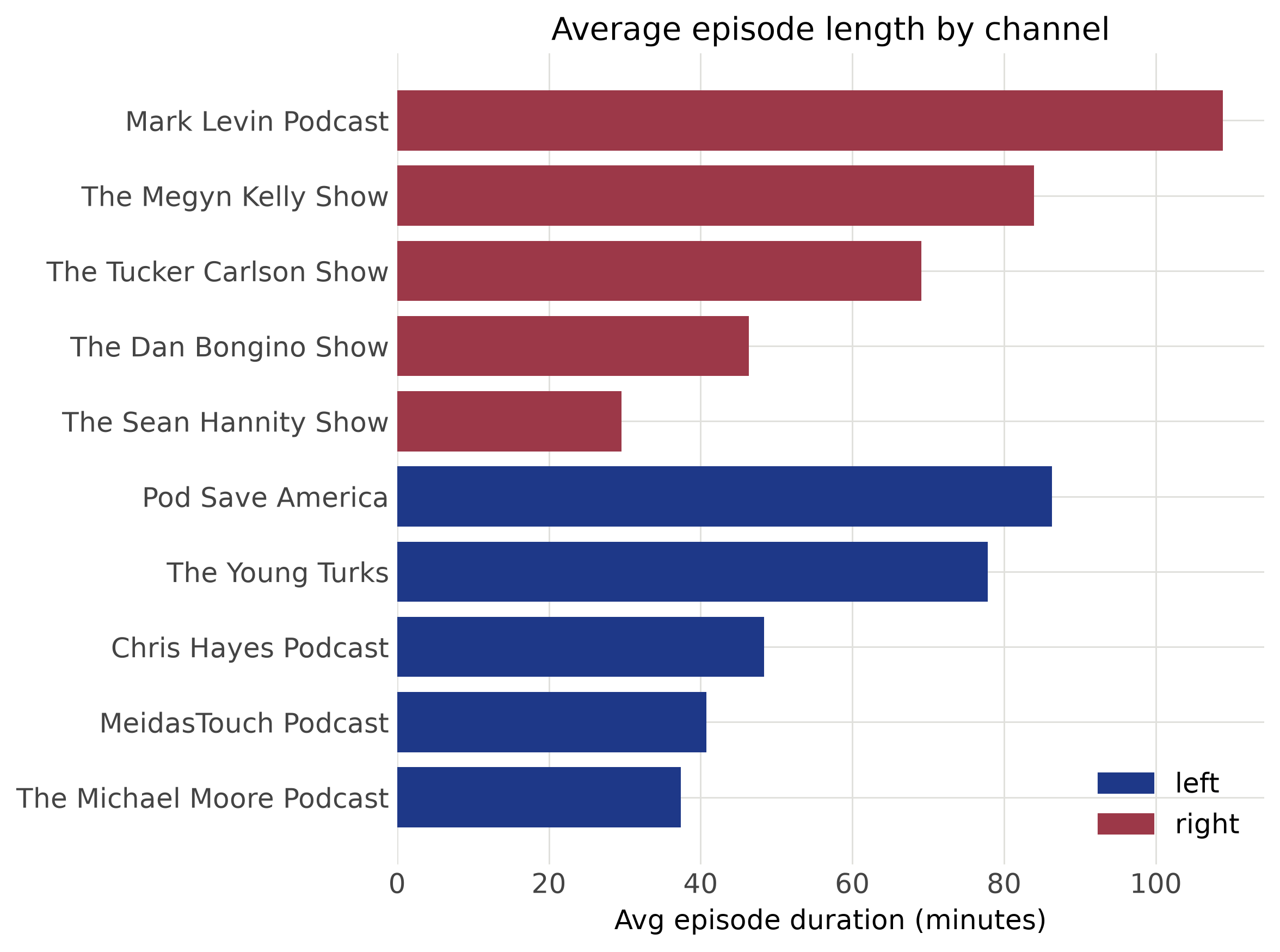}
    \caption{Average episode duration by channel.}
    \label{fig:avg-duration}
\end{figure}

The final list of channels used for analysis is \say{Pod Save America}, \say{The MeidasTouch Podcast}, \say{The Young Turks}, \say{Why is this happening?} (The Chris Hayes Podcast), and \say{The Michael Moore Podcast} on the left and \say{The Dan Bongino Show}, \say{The Tucker Carlson Show}, \say{The Megyn Kelly Show}, \say{The Sean Hannity Show}, and the \say{Mark Levin Podcast} on the right. A summary of the coverage per channel-year is included in Fig. \ref{fig:cov-channel-year} and the number of sampled linguistic metaphors per channel-year is shown in Fig. \ref{fig:mets-channel-year}. We show the average episode length for each channel in Fig. \ref{fig:avg-duration}.

\subsection{Additional Results}
Here we continue our analysis on the metaphor in podcasts about U.S. political podcasts from Sec. \ref{sec:case-study}, thus demonstrating the capabilities of our conceptual metaphor detection framework. 

\begin{table*}[t]
\centering
\footnotesize
\setlength{\tabcolsep}{3pt}
\renewcommand{\arraystretch}{1.15}
\begin{tabular}{@{}c p{0.28\linewidth} p{0.52\linewidth}@{}}
\toprule
\textbf{Rank} & \textbf{Conceptual Metaphor} & \textbf{Linguistic Metaphors} \\
\midrule
1 &
\textbf{Lawmakers are objects under external control.} &
``...a group of hardliners were trying to \textit{pressure} Johnson to only put Israel aid on the floor and hold Ukraine aid until the Senate passes the border bill...''
``...do they \textit{control} Pelosi and Biden and Schumer?''
``...they're the ones who pressured bill buckley...'' \\
\midrule
3 &
\textbf{The economy is a system subject to vertical changes.} &
``...because it's not \textit{going down} dramatically and reverting back to old prices...''
``...why is electricity spiking?''
``Even Disneyland is \textit{raising} prices on food, and they're blaming inflation...''
``The White House is hailing the average wholesale price of a dozen eggs has dropped 47 percent since they took office...'' \\
\midrule
19 &
\textbf{Information, power, and media stories are a weapon.} &
``...they will use [information] without any hesitance to \textit{crush} your life...''
``...trying to \textit{wield} power over the rest of us based on that demented psychology.''
``...trying to \textit{build} a story that is a powerful narrative of who's the good guy and who's the bad guy''
``Disney's \textit{pushing} that [narrative].''
``...to use a pandemic to \textit{seize} control of people's freedom and their money.'' \\
\midrule
25 &
\textbf{Immigration as an active invasion/crisis being fought.} &
``That is priority one, getting \textit{control} of the immigration crisis, the illegal invasion of our country...''
``Drug cartels \textit{control} our border.''
``...these are two of the centers that have been \textit{overwhelmed}... with these migrants that have been shipped there by the Biden administration.'' \\
\bottomrule
\end{tabular}
\caption{Ranked conceptual metaphor clusters and representative linguistic instantiations.}
\label{tab:cs-qual-analysis}
\end{table*}

\begin{figure}
    \centering
    \includegraphics[width=1\linewidth]{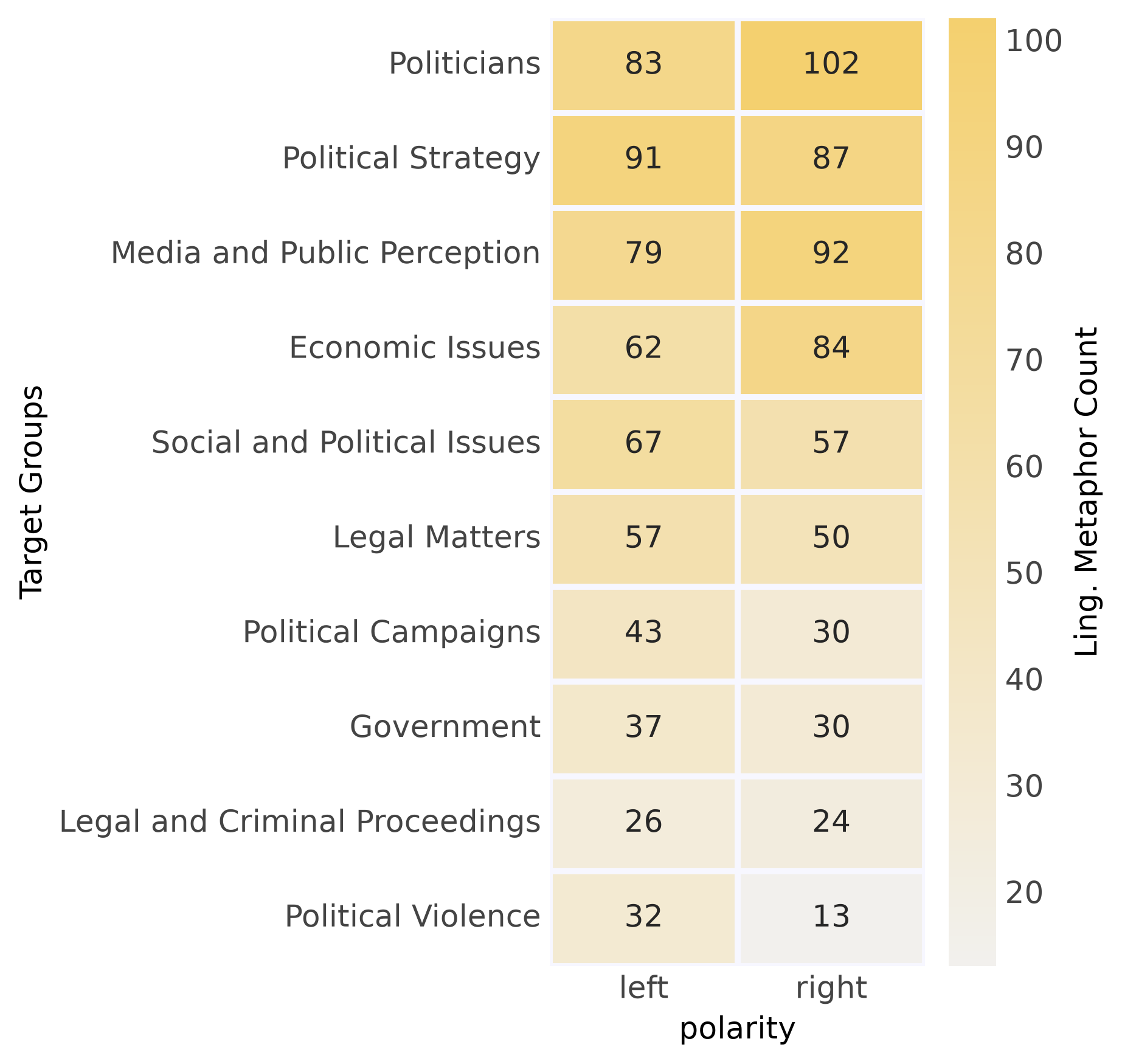}
    \caption{Most commonly-used target groups by the left vs. right.}
    \label{fig:tg-heatmap}
\end{figure}

\paragraph{Are certain target groups more common in left- or right-leaning podcasts?}
Fig. \ref{fig:tg-heatmap} shows that the most common target groups appearing in linguistic metaphors across the dataset are Politicians, Political Strategy, Media and Public Perception, and Economic Issues. However, these target groups are also among the most divisive (Fig. \ref{fig:div-tg}. Right-leaning sources are significantly more likely to metaphorically characterize Economic issues and Politicians, while left-leaning sources are more likely to include Political Violence and Political Campaigns. 


\section{Prompts}
\label{app:prompts}
The prompt for metaphor classification is shown in Prompt \ref{prompt:metaphor_binary_classification}. The prompt for generating metaphor implications is shown in Prompt \ref{prompt:gen_entailments}. The prompts for identifying salient target groups and assigning target groups to target nouns are shown in Prompt \ref{prompt:create-target-groups}, Prompt \ref{prompt:clean-target-groups}, and Prompt \ref{prompt:target_group_ann}. The prompt for assigning image schema groups to source verbs is in Prompt \ref{prompt:source_group_ann}. 

\section{Codebooks}
\label{app:codebooks}
The codebook for evaluating linguistic metaphor interpretations is included in Fig. \ref{cb:ling-met-rep}.

\section{Target Groups}
\label{app:tg-tabs}
The generated target groups and their accompanying definitions are included for the immigration corpus in Tab. \ref{tab:imm-tg}, gun control in Tab. \ref{tab:gc-tg}, abortion in Tab. \ref{tab:a-tg}, and podcasts in Tab. \ref{tab:p-tg}.

\section{License For Pre-Existing Artifacts}
All pre-existing artifacts utilized are publicly available under open-source and open-access licenses. This work adheres to all intended use guidelines and terms of service for the respective resources.

\section{Compute}
LLM inference was run on Nvidia H100 and A100 1485 GPUs. Clustering experiments were conducted on either CPU or a single 1487 Nvidia A100. Pipeline execution requires 5-10 days depending on the dataset.



\onecolumn
\small
\begin{longtable}{
  >{\raggedright\arraybackslash}p{0.12\linewidth}
  >{\raggedright\arraybackslash}p{0.10\linewidth}
  >{\raggedright\arraybackslash}p{0.42\linewidth}
  >{\raggedright\arraybackslash}p{0.25\linewidth}
}
\caption{Summaries for the top clusters achieved by each ablation setting for the immigration corpus.}
\label{tab:ablation-qual-summary} \\
\toprule
\textbf{Clustering Method} & \textbf{Cannot-link Constraints} & \textbf{Clusters} & \textbf{Summary} \\
\midrule
\endfirsthead

\multicolumn{4}{c}{\tablename\ \thetable{} -- continued from previous page} \\
\toprule
\textbf{Clustering Method} & \textbf{Cannot-link Constraints} & \textbf{Clusters} & \textbf{Summary} \\
\midrule
\endhead

\midrule
\multicolumn{4}{r}{\textit{Continued on next page}} \\
\endfoot

\bottomrule
\endlastfoot

kmeans (ling. met only) & --- &
(1) sentences criticizing judges in different contexts.\newline\newline
(2) no topical or metaphorical consistency \newline\newline
(3) conversation of women in different contexts.
&
Irregularly-sized clusters: important ones are very small with high semantic similarity, high-level topical consistency (e.g., \textit{women}) \\
\midrule

kmeans (textual met. interpretation) & --- &
(1) repair broken object (\textit{fix, solve}) applied to problems and systems with a few stragglers (\textit{negotiate}, \textit{refresh}); largely bipartisan concerns about failing system, calls for repair \newline\newline
(2) various source domains (\textit{burden, drop, flood, ruined}) applied to immigrants; emphasizes the toll of immigration on the system \newline\newline
(3) no metaphorical or framing consistency, broad discussion of heritage &
Strong topical consistency across clusters but fails to distinguish between different source--target pairs \\
\midrule

pckmeans & image schema group &
(1) no metaphorical consistency (excepting repeated use of `give') no framing consistency, broad discussion of the media \newline\newline
(2) no similarity across metaphorical info (breaks, feed, licking, made); similar topically (Fox news, Trump, Trump supporters) with consistently left-leaning spin on events \newline\newline
(3) physical object applied to ideas; similar topically but inconsistent framing (immigrant vs. natural citizen well-being) &
The image schema group cannot-link constraint weakens the topical and/or framing consistency of K-means alone. Vague similarity between metaphorical verbs but doesn't distinguish between source domains. \\
\\
\\
 & target group &
(1) single linguistic metaphor \newline\newline
(2) no metaphorical consistency; discusses separating children from their families with mostly anti-immigration framing \newline\newline
(3) various source domains (flooding, hurt, create, inundated) applied to immigration; consistent `harms of immigration' framing &
Improved topical consistency due to cannot-link target group constraint, some framing consistency, but with sprawling source domains. \\
\\
\\
 & image schema group, noun type &
(1) border as container (open, shuts, seal, stopping), consistent ``unsecured border'' framing. \newline\newline
(2) citizenship/benefits-as-reward (rewarded, anchor, giving, have perks, provide), consistent ``undeserved benefits'' for immigrants framing \newline\newline
(3) no metaphorical or framing consistency; broad discussion of language and communication &
Improved metaphorical consistency with less-specific topical focus. \\
\\
\\
 & image schema group, target group &
(1) spatial motion-based source domains (putting, keep) applied to economic power, consistent ``keeping money at the top'' applied to ``corporations exploiting U.S. workers'' framing \newline\newline
(2) sprawling source domains applied to the target domain of profiting off labor, consistent framing of Donald Trump as profiting off of immigrant labor \newline\newline
(3) frames right-leaning rhetoric around immigration as ``fear-mongering'' &
Most specific topical and framing consistency. Similar verb usage allows for high-quality qualitative analysis of the metaphorical source domains used to achieve the specific framing. \\
\end{longtable}

\twocolumn
\begin{table*}[t]
\centering
\footnotesize
\renewcommand{\arraystretch}{0.95}
\setlength{\tabcolsep}{5pt}
\begin{tabular}{
  >{\raggedright\arraybackslash}p{0.09\linewidth}
  >{\raggedright\arraybackslash}p{0.28\linewidth}
  >{\raggedright\arraybackslash}p{0.55\linewidth}
}
\toprule
\textbf{Importance Ranking} & \textbf{Framing} & \textbf{Linguistic Metaphors} \\
\midrule

1 &
\textbf{Gun ownership qualification is entering a building.} Passing background checks required for gun ownership as entering a building. The examples frame background checks as a defense for a physical structure to be strengthened to prevent entering (owning a gun) easily. &
``John Cornyn, R-Texas and Chris Murphy, D-Conn., to \textit{strengthen} FBI background checks'' \newline
``He has previously expressed an interest in efforts to \textit{strengthen} the federal background check system.'' \newline
``Steps to \textit{strengthen} background checks could come this week.'' \newline
``...tell them to \textit{close} background check loopholes\ldots'' \newline
``Reid plans to bring a watered-down gun control bill that includes \textit{expanded} background checks\ldots'' \\
\midrule

2 &
\textbf{Political disagreements are a physical struggle/war.} Bipartisan sources discussing political debate, rhetoric, and legislation &
``Republicans \textit{pushed back} on the notion\ldots'' \newline
``Democrats \textit{seized} the moment\ldots'' \newline
``Jimmy Kimmel came under fire after \textit{blasting} Republicans'' \newline
``Republicans fiercely \textit{resisted} the Democratic pressure\ldots'' \newline
``...if the Republican Senate \textit{quashes} Democrats' gun schemes'' \newline
``...\textit{destroy} the GOP\ldots'' \newline
``...if Trump is reelected, if the Republicans \textit{hold} the Senate, and if they re-take the House of Representatives\ldots'' \\
\midrule

4 &
\textbf{Gun ownership rights are an object on a scale.} U.S. constitutional second amendment rights are compared to an object on a scale. Public safety is on the other side of the scale and the goal is to balance the two by regulating rather than banning gun ownership. &
``\textit{Balancing} the rights of those with mental-health issues and the desire to safeguard the public\ldots'' \newline
``...I think the risk outweighs the benefits.'' \newline
``...with one gun owner remarking that his `convenience doesn't \textit{outweigh} the risk of use' of such weapons.'' \newline
``the bill\ldots \textit{balances} `our individual rights with need for public safety.''' \\

\bottomrule
\end{tabular}
\caption{Qualitative analysis of top-ranked metaphor clusters by importance (gun control corpus).}
\label{tab:gc-qual-analysis}
\end{table*}

\begin{table*}[t]
\centering
\footnotesize
\renewcommand{\arraystretch}{0.95}
\setlength{\tabcolsep}{5pt}
\begin{tabular}{
  >{\raggedright\arraybackslash}p{0.09\linewidth}
  >{\raggedright\arraybackslash}p{0.28\linewidth}
  >{\raggedright\arraybackslash}p{0.55\linewidth}
}
\toprule
\textbf{Importance Ranking} & \textbf{Framing} & \textbf{Linguistic Metaphors} \\
\midrule

1 &
\textbf{Opinions/stances are physical structures.} Bipartisan discussion of abortion stances as a structure such as a building that is made up of and supported by physical materials (people aligned with the stance). &
``...prevents the Republican Party from taking advantage of widely \textit{supported} pro-life positions\ldots'' \newline
``...voting for candidates who \textit{supported} abortion rights `opened the door to the curse.''' \newline
``Conservative Christians who \textit{form} the bulk of the anti-abortion movement\ldots'' \newline
``Dan Lipinski, who has been vocal about \textit{maintaining} an anti-abortion stance\ldots'' \\
\midrule

3 &
\textbf{Science and pro-choice stances are competing destructive forces.} The examples mostly frame modern science as a force to destroy pro-choice stances by acknowledging abortion as death. &
``...refusing to acknowledge abortion as a death \textit{undermines} the role of science.'' \newline
``Modern science likewise \textit{demolishes} the dismissive `clump of cells' justification that abortion advocates of the 1970s told themselves to soothe tender consciences.'' \newline
``...modern medical science is \textit{forcing} abortion supporters to admit the grisly reality of what they advocate\ldots'' \newline
``Science does not bear either assumption out.'' \newline
``When we ignore biological facts, social arrangements tend to denigrate biological difference\ldots'' \\
\midrule

5 &
\textbf{Media attention or stories are physical objects.} The examples frame news stories as physical objects to be caught, revealed, or buried. &
``Let's hope the news media catch on to Paul Ryan.'' \newline
``...an old video of the counselor to President Donald Trump threatening to `perform' abortions on feminists using guns was \textit{resurfaced} by a well-known media news outlet.'' \newline
``James Hodgkinson's ties to the Bernie Sanders campaign have been reported (if sometimes \textit{buried}) by most news outlets\ldots'' \newline
``...and we see them get press during MLK Day\ldots'' \newline
``Mediaite on Monday \textit{resurfaced} a 2007 speech Conway made\ldots'' \\
\midrule

13 &
\textbf{Abortion is a physically destructive force.} The examples frame abortion as a destructive force (physical attack or disease) disproportionately affecting minority groups from a pro-life perspective. &
``...there is an additional national tragedy and injustice when the scourge of abortion especially \textit{ravages} racial or ethnic groups who also face centuries-old racial discrimination and segregation.'' \newline
``It is clear that a large number of black American lives are being \textit{taken} by abortions each year.'' \newline
``...the argument that abortion \textit{poses} a unique threat to black lives has seen an increase in attention in recent years\ldots'' \newline
``They're \textit{facing} a lot of the brunt of this and it's further perpetuating disparities.'' \\

\bottomrule
\end{tabular}
\caption{Qualitative analysis of top-ranked metaphor clusters by importance (abortion corpus).}
\label{tab:abor-qual-analysis}
\end{table*}

\onecolumn
\begin{longtable}{
  >{\raggedright\arraybackslash}p{0.11\linewidth}
  >{\raggedright\arraybackslash}p{0.20\linewidth}
  >{\raggedright\arraybackslash}p{0.14\linewidth}
  >{\raggedright\arraybackslash}p{0.20\linewidth}
  >{\raggedright\arraybackslash}p{0.20\linewidth}
}
\caption{Examples of LLM annotation errors by error type.}
\label{tab:error-examples} \\
\toprule
\textbf{Error Type} & \textbf{Context Sentence} & \textbf{Target / Source Information} & \textbf{Frame Implications} & \textbf{Error Explanation} \\
\midrule
\endfirsthead

\multicolumn{5}{c}{\tablename\ \thetable{} -- continued from previous page} \\
\toprule
\textbf{Error Type} & \textbf{Context Sentence} & \textbf{Target / Source Information} & \textbf{Frame Implications} & \textbf{Error Explanation} \\
\midrule
\endhead

\midrule
\multicolumn{5}{r}{\textit{Continued on next page}} \\
\endfoot

\bottomrule
\endlastfoot

Incorrect metaphor candidate extraction &
On Tuesday, a dispute between two people at Lone Star College in Houston Texas generally has lax gun laws, but does your state allow concealed guns on college campuses? &
\textbf{dispute:} Thing, Other \newline\newline \textbf{has:} Other &
The state's possession of lax gun laws is causing disorder on college campuses. The state's failure to enforce strict gun control has led to unsafe conditions. It is irresponsible for the state to allow lax gun laws that endanger students. The state should eliminate lax gun laws and implement stricter regulations. &
The shortest dependency path between the specified target noun and source verb does not match any of the pre-determined paths associated with a metaphor. \\
\midrule

Incorrect metaphor classification &
Second, studies of masculinities and men's health suggest patriarchal social systems can foster a toxic culture that harms men as well as women. &
that: Thing, Other \newline\newline \textbf{harms}: Other &
Patriarchal social systems create a toxic culture that negatively affects both men and women. Patriarchal social systems foster a toxic environment that leads to harm. It is unethical that patriarchal systems cause harm to both genders. Patriarchal social systems should be reformed to eliminate the toxic culture causing harm. &
Because the context sentence is describing a \textit{physical harm} to men due to patriarchal social systems, the verb is not being used metaphorically. \\
\midrule

Incorrect target group classification &
Many speculated that de Blasio's victory could herald a new national trend. &
victory: Thing, \textbf{Political Entities and Advocacy Groups} \newline\newline herald: Other &
A significant national shift is emerging due to de Blasio's victory. De Blasio's victory acts as a signal that the national trend is changing. The new national trend is viewed as a positive development. Support and accelerate the new national trend indicated by de Blasio's victory. &
The noun classification is correct, indicating that the LLM initially focused on the abstract noun `victory'. However, when prompted for a target group classification, the LLM focused on the more concrete political figure `de Blasio'. \\
\midrule

Incorrect image schema group classification &
Obama submitted an ambitious gun-control plan to Congress yesterday, which called for a ban on assault weapons and large ammo clips as well as expanded background checks for gun buyers. &
checks: Thing, Gun Control Policies and Advocacy \newline\newline \textbf{expanded:} Other &
The current background checks are insufficient to prevent gun violence. The government's initiative to increase the scope of background checks is necessary to address gun violence. Expanding background checks is a morally responsible action to enhance public safety. The government should fully implement the expanded background checks as part of the gun-control plan. &
Rather than `other', the image schema group should have been labeled `spatial motion'. \\
\midrule

Incorrect frame implication generation &
The newly drawn 10th District in southern Pennsylvania still leans Republican, but not as heavily as before. &
district: Place, Congressional Districts \newline\newline drawn: Spatial Motion &
\textbf{The 10th District's political boundaries have been manipulated to shift its lean toward Republican. The district's boundaries were intentionally redrawn to alter its political outcome. It is unethical to manipulate electoral boundaries for political advantage.} The district's boundaries should be redrawn fairly without partisan influence. &
The sentence more likely implies that the district boundary was manipulated to lean Democrat, rather than Republican. In spite of this error, the frame implications are relevant to the metaphor invoked by the verb "drawn." \\
\midrule

Irrelevant frame implication generation &
``these are animals''??wtf THESE ARE HUMAN BEINGS WHO WANT A BETTER LIFE IN A OUR COUNTRY . he fails to realize immigrants come here for more opportunities than they have in their own country we should feel HONORED that they want to come here. &
we: Person, U.S. Citizens \newline\newline feel: Other &
\textbf{People are dehumanizing immigrants by treating them as animals rather than humans. Immigrants are seeking better opportunities, which the speaker believes is being ignored or misunderstood. It is immoral to reduce immigrants to animals and dismiss their human aspirations.} People should feel honored by immigrants' desire to seek a better life in the country. &
In spite of the implications being consistent with the sentence, they are not relevant to the identified metaphor, instead focusing on different parts of the original sentence. \\

\end{longtable}

\normalsize
\twocolumn


\input{8_prompts}


\input{9_codebooks}


\input{tg_tables}

%% file: 8_prompts.tex
\clearpage

\renewcommand{\thetextbox}{\thesection.\arabic{textbox}}

\begin{textbox*}[htbp]
\centering
\begin{tcolorbox}[
    colback=lightgreen,            
    colframe=darkgreen,            
    width=\textwidth,              
    arc=3mm,                       
    boxrule=1pt,                   
    left=5mm,                      
    right=5mm,                     
    top=3mm,                       
    bottom=3mm                     
]
{\ttfamily\small

\textbf{System Prompt}\\
\\
Instruction: You are an annotator who is developing a dataset for measuring metaphors. Your response should be in JSON format with the key `classification' and the value Literal or Metaphorical. Additionally, in the JSON, you should indicate with the key `explanation' the justifications for your decision. If you select `Metaphorical', then your explanation should describe which metaphorical aspects from the source verb are being used to characterize the target word. \\
\\
Desired JSON: \{`classification': str, `explanation': `your reasoning for the answer'\}.\\
\\
Do not generate anything else.\\
\\
\\
\textbf{User Prompt}\\
\\
Question: Analyze the use of the specified source verb in the sentence provided. Focus only on determining whether this verb, in its specific use within the sentence, is used literally, that is, describing the physical action, or whether it is used metaphorically, where the use of the verb transcends its original meaning without referring directly to a physical action, such as, for example, giving some kind of personification or animalization of the target noun. It is important to distinguish the specific lexical analysis of the verb from any broader metaphorical interpretation that may arise from comparisons or conceptual equivalences present in the sentence. Please provide your evaluation focusing solely on the specified verb and how it is used to characterize the target noun.\\
\\

}
\end{tcolorbox}
\caption{Binary metaphor classification.}
\label{prompt:metaphor_binary_classification}
\end{textbox*}

\clearpage

\renewcommand{\thetextbox}{\thesection.\arabic{textbox}}

\begin{textbox*}[htbp]
\centering
\begin{tcolorbox}[
    colback=lightgreen,            
    colframe=darkgreen,            
    width=\textwidth,              
    arc=3mm,                       
    boxrule=1pt,                   
    left=5mm,                      
    right=5mm,                     
    top=3mm,                       
    bottom=3mm                     
]
{\ttfamily\small

\textbf{System Prompt}\\
\\
Instruction: Human utterances communicate propositions that may or may not be explicit in the literal meaning of the utterance. Metaphorical language often contains implicit entailments guided by the logic of the metaphor. For each utterance, state 4 implicit propositions communicated by the specified metaphorical verb. The 4 propositions should answer the following questions: (1) What is the problem definition implied by the author's metaphor choice? (2) What is the implied cause of the problem? (3) What is the author's moral evaluation of this situation? (4) What is the implied treatment recommendation? Implicit propositions may be inferences about the subject of the utterance or about the perspective of its author. All generated propositions should be short, independent, and written in direct speech and simple sentences. If possible, write propositions with a subject, verb, and object.\\
\\
Desired JSON: \{`problem\_def': `', `cause': `', `moral\_eval': `', `treatment\_rec': `'\}.\\
\\
Do not generate anything else.\\
\\
\\
\textbf{User Prompt}\\
\\
Question: List the implicit propositions for the given sentence. Focus only on the entailments rising from the metaphorical verb. \\
\\
\textbf{Example 1}\\
Metaphorical verb: `flooding'\\
Sentence: `Tell Congress Louisville, Kentucky is a border town due to Democrat mayor flooding illegal aliens into our city.'\\
\\
Answer: \{\\
    `problem\_def': `There has been an increase in immigration, a destructive force, in the city.',\\
    `cause': `The democrat mayor has allowed immigrants to enter the city.',\\
    `moral\_eval': `It is immoral that the mayor would allow the immigrants to destroy the city.',\\
    `treatment\_rec': `The \say{flow} of immigration into the city should be stopped.'\\
\}\\
\\
\textbf{Example 2}\\
Metaphorical verb: `caging'\\ 
Sentence: `Trump admin :   - Destroying the environment   - Caging children  - targeting legal migrants   - Racism   - packing federal courts with radicals'\\
\\
Answer: \{\\
    `problem\_def': `The Trump administration is imprisoning immigrant children.',\\
    `cause': `The Trump administration does not treat immigrant children with human decency, as they \say{cage} them like animals.',\\
    `moral\_eval': `It is immoral that the Trump administration would treat immigrant children as animals rather than people.',\\
    `treatment\_rec': `The Trump administration should free the immigrant children rather than continue to imprison them.'\\
\}\\

}
\end{tcolorbox}
\caption{Generate metaphor entailments guided by \citet{entmanFramingClarificationFractured1993}'s framing effects. Adapted from \citet{hoyleNaturalLanguageDecompositions2023}'s prompts for generating implicit entailments from utterances.}
\label{prompt:gen_entailments}
\end{textbox*}

\clearpage

\renewcommand{\thetextbox}{\thesection.\arabic{textbox}}

\begin{textbox*}[htbp]
\centering
\begin{tcolorbox}[
    colback=lightgreen,            
    colframe=darkgreen,            
    width=\textwidth,              
    arc=3mm,                       
    boxrule=1pt,                   
    left=5mm,                      
    right=5mm,                     
    top=3mm,                       
    bottom=3mm                     
]
{\ttfamily\small

\textbf{System Prompt}\\
\\
Instruction: You are an expert researcher who wishes to understand which [NOUN\_GROUP]s are referenced in a particular dataset with respect to a particular issue. Your task is to determine clean [NOUN\_GROUP] group label(s) that exist in the set of nouns given their context. \\
\\
Desired JSON: \{`groups': [groups in dataset]\}.\\
\\
Do not generate anything else.\\
\\
\\
\textbf{User Prompt}\\
\\
Question: What are some [NOUN\_GROUP]s referenced below? The labels you produce should be general, capturing relatively broad organizations that are important in a discourse. However, they should not be so general that they don't capture anything meaningful with respect to framing. The labels should capture no bias or judgment. For example 'democratic party' is preferred over `extremist democratic party'.\\

}
\end{tcolorbox}
\caption{Generating target group labels.}
\label{prompt:create-target-groups}
\end{textbox*}

\clearpage

\renewcommand{\thetextbox}{\thesection.\arabic{textbox}}

\begin{textbox*}[htbp]
\centering
\begin{tcolorbox}[
    colback=lightgreen,            
    colframe=darkgreen,            
    width=\textwidth,              
    arc=3mm,                       
    boxrule=1pt,                   
    left=5mm,                      
    right=5mm,                     
    top=3mm,                       
    bottom=3mm                     
]
{\ttfamily\small

\textbf{System Prompt}\\
\\
Instruction: You are an expert researcher who wishes to understand groups of people/places/things referenced in a particular dataset with respect to a particular issue. Your task is to clean a provided set of group labels and corresponding definitions by combining as many as possible and removing biased words. The definitions should be descriptive and provide an understanding of the group's role with respect to the issue.\\
\\
Desired JSON: \{`group\_label': `group definition', ...\}.\\
\\
Do not generate anything else.\\
\\
\\
\textbf{User Prompt}\\
\\
Question: Construct a clean list of group labels by combining ones that are provided. No group should be a subset of another. Groups should remain separate if the distinction is important for the framing of a particular issue in the U.S. For example, `undocumented immigrants' and `documented immigrants' should remain separate, as this distinction has implications for framing immigration issues. Alternatively, `liberals' and `democrats' may be combined. Remove group labels that carry bias or judgment such as `freeloaders'. Also remove overly generic groups such as `countries' or `everyone' which provide no understanding of bias in the dataset.\\

}
\end{tcolorbox}
\caption{Consolidating target group labels.}
\label{prompt:clean-target-groups}
\end{textbox*}

\clearpage

\renewcommand{\thetextbox}{\thesection.\arabic{textbox}}

\begin{textbox*}[htbp]
\centering
\begin{tcolorbox}[
    colback=lightgreen,            
    colframe=darkgreen,            
    width=\textwidth,              
    arc=3mm,                       
    boxrule=1pt,                   
    left=5mm,                      
    right=5mm,                     
    top=3mm,                       
    bottom=3mm                     
]
{\ttfamily\small

\textbf{System Prompt}\\
\\
Instruction:  You are a grammar expert building a dataset for noun classification. Your task is to determine whether the target noun is a person (or group of people), organization, place, or thing. \\
\\
Desired JSON: \{`noun\_classification': `[person/place/thing/organization]'\}.\\
\\
Do not generate anything else.\\
\\
\\
\textbf{User Prompt}\\
\\
Question: Is the the target noun specified below a person (or group of people), organization, place or thing? You may use the context sentence to help make your decision, but provide your answer focusing solely on the specified noun.\\
\\

}
\end{tcolorbox}
\caption{Target noun classification annotation.}
\label{prompt:target_noun_class_ann}
\end{textbox*}

\clearpage

\renewcommand{\thetextbox}{\thesection.\arabic{textbox}}

\begin{textbox*}[htbp]
\centering
\begin{tcolorbox}[
    colback=lightgreen,            
    colframe=darkgreen,            
    width=\textwidth,              
    arc=3mm,                       
    boxrule=1pt,                   
    left=5mm,                      
    right=5mm,                     
    top=3mm,                       
    bottom=3mm                     
]
{\ttfamily\small

\textbf{System Prompt}\\
\\
Instruction:  You are an annotator who is developing a dataset for analyzing U.S. political podcast transcripts. Your goal is to map nouns referring to [NOUN\_GROUP]s to broader group labels. For each submission, you will identify the target group that matches the specified target [NOUN\_GROUP]. \\

Group Assignment:\\
- Map the identified target noun to the most appropriate group.\\
- If no group from the provided list adequately represents the target, assign it to 'Other'. This includes entities that are tangentially related but don't fit well into any specific predefined category.\\
- If the correct target group assignment is not clear from the given context, assign it to `Unclear'\\
- If the noun was misclassified as a [NOUN\_GROUP], assign it to `Not a [NOUN\_GROUP]'\\
\\
Each annotation task will include:\\
GROUPS: List of predefined group categories relevant to the domain\\
TARGET NOUN: the organization noun we wish to categorize\\
CONTEXT SENTENCE: the full sentence from the podcast transcript containing the target noun.\\
\\
Desired JSON: \{`target\_group': `[Identified group from predefined list]'\}\\
\\
Do not generate anything else.\\
\\
\\
\textbf{User Prompt}\\
\\
Question: Please provide a group label focusing only on the noun specified.\\
\\

}
\end{tcolorbox}
\caption{Target group annotation.}
\label{prompt:target_group_ann}
\end{textbox*}

\clearpage

\renewcommand{\thetextbox}{\thesection.\arabic{textbox}}

\begin{textbox*}[htbp]
\centering
\begin{tcolorbox}[
    colback=lightgreen,            
    colframe=darkgreen,            
    width=\textwidth,              
    arc=3mm,                       
    boxrule=1pt,                   
    left=5mm,                      
    right=5mm,                     
    top=3mm,                       
    bottom=3mm                     
]
{\ttfamily\small

\textbf{System Prompt}\\
\\
Instruction:  You are a metaphor expert building a dataset for characterizing metaphorical source domains. Your task is to determine what image schema group a metaphor invoked by a given verb falls into. Some examples of metaphorical uses of the verb are provided.\\
\\
Desired JSON: \{`image\_schema\_group': `[spatial motion/force/balance/other]'\}\\
\\
Do not generate anything else.\\
\\
\\
\textbf{User Prompt}\\
\\
Question: Based on the examples below, what kind of image schema does the source verb invoke? Choose 'spatial motion', 'force', 'balance' or 'other'. Do not generate anything else.\\
\\

}
\end{tcolorbox}
\caption{Source image schema group annotation.}
\label{prompt:source_group_ann}
\end{textbox*}

%% file: 9_codebooks.tex



\clearpage
\onecolumn

\refstepcounter{table}
\addcontentsline{lot}{table}{\protect\numberline{\thetable}{Codebook for assessing quality of linguistic metaphor representations.}}
\label{cb:ling-met-rep}

\begin{tcolorbox}[
    colback=lightblue0,             
    colframe=darkblue0,             
    breakable,                     
    width=\textwidth,              
    arc=3mm,                       
    boxrule=1pt,                   
    left=5mm,                      
    right=5mm,                     
    top=3mm,                       
    bottom=3mm                     
]
{\small

\begin{center}\textbf{\Large Instructions}\end{center}

\medskip
\noindent\textbf{Column Label Key}

\smallskip
\noindent
\texttt{correct\_ng} = correct noun group\\
\texttt{correct\_tg} = correct target group\\
\texttt{correct\_ss} = correct source schema

\medskip
\noindent ** Content column names are shown in blue and columns to be annotated are in red.

\smallskip
\noindent ** Note that throughout, the word \textit{plausible} refers to whether a reasonable reader would accept the generated annotations without needing to stretch the definition, even if a better label or generated text exists.

\bigskip
\noindent\textbf{Example Linguistic Metaphors}

\smallskip
\noindent
Taxes \textit{burden} the middle class.\\
The middle class shouldn't \textit{carry} the bulk of taxation.

\bigskip
\noindent\textbf{\large Instructions}

\begin{enumerate}
    \item \textbf{Start by identifying the metaphor in the `sentence' column.}
    \begin{enumerate}
        \item The word in the \texttt{source\_word} column is the metaphorical verb we're focusing on. A verb is being used metaphorically if its use transcends its original meaning without referring directly to a physical action.
        \item The \texttt{target\_word} is the noun being characterized by the metaphor. Identify both in the sentence.
        \item Note that you may not agree that the \texttt{source\_word} is metaphorical. If this is the case, you will mark the \texttt{metaphor\_implications} column as `Not Metaphorical'. Continue with the other annotations; the accuracy of the metaphor classification component is evaluated in a different step.
    \end{enumerate}

    \item \textbf{Complete annotations related to the \texttt{target\_word}.} This includes the \texttt{correct\_ng} and \texttt{correct\_tg} columns. The goal here is to identify broader groups that the target noun belongs in. The steps are included below, as well as on the `Target Word' tab of this document.
    \begin{enumerate}
        \item Identify the target word in the context sentence.
        \item Determine if the target word has been correctly defined as person/place/thing/organization and mark \texttt{correct\_ng} accordingly. If it has been correctly defined, continue to annotate \texttt{correct\_tg}. If not, leave \texttt{correct\_tg} blank.
        \begin{enumerate}
            \item Note that there may be some overlap in the person/thing/organization labels. If the classification is plausible, mark it as correct.
        \end{enumerate}
        \item Navigate to the subtab of this document tab corresponding to the dataset you're working on. This is the name of the dataset and will be included as the title of the annotation sheet you're working on. The tab will show the list of possible target groups for each noun group. Navigate to the appropriate noun group on the tab. These are the possible labels for the target word's target group. Determine whether the target group classification is plausible for the target noun; mark \texttt{correct\_tg} accordingly.
    \end{enumerate}

    \item \textbf{Complete annotation related to the \texttt{source\_word}.} This includes the \texttt{correct\_ss} column. The source image schema group is meant to help characterize the source domain of the metaphor. The goal is to match the metaphorical source verb to the group that best describes the action that the verb would describe if used literally. Note that if you do not agree that the source word is being used metaphorically, you may still annotate this column based on the literal action the verb describes. Details on the groups are included in the `Image Schema Groups' tab.
    \begin{enumerate}
        \item Identify the source word in the sentence.
        \item Determine what action the verb would describe if it were used literally.
        \item Does this action plausibly fit into the category determined by the LLM?
        \item The categories and subcategories are not mutually exclusive. In other words, a verb can fit into multiple categories. If the LLM-assigned image schema group is plausible, mark True.
    \end{enumerate}

    \item \textbf{Complete annotations related to the \texttt{frame\_implications}.} This includes the \texttt{plausible\_implications} and \texttt{metaphor\_implications}.
    \begin{enumerate}
        \item First determine whether \textit{all} the sentences listed are plausible implications of the sentence as a whole. If they are, mark \texttt{plausible\_implications} as true. If even one is implausible, mark false and leave \texttt{metaphor\_implications} blank.
        \item If you do not agree that the source word is used metaphorically, leave the \texttt{metaphor\_implications} column blank. Otherwise, determine whether the sentences are specifically related to the metaphor. Sentences will be related to the metaphor if they are guided by the logic of the metaphor. In the \textbf{Example Metaphors} above, the implications must be drawn from the logic of carrying taxes like a heavy burden; it is immoral for less able people to carry it, taxes should be lifted from peoples' shoulders, etc. If they do follow from this logic, mark \texttt{metaphor\_implications} as true. Once again, if even one doesn't plausibly relate to the metaphor, mark false.
    \end{enumerate}
\end{enumerate}

\noindent For all columns, there is also an `Unsure' option. This may be used as a marker to think about and come back to annotations. It may also be used to communicate that there is not enough information given the context to make a decision.

\bigskip
\noindent\textbf{\large Example 1}

\smallskip
\textbf{sentence:} Isn’t marrying your brother to skirt our immigration system a crime?\\
\textbf{target word:} crime\\
\textbf{noun group:} thing\\
\textbf{target group:} Anti-Immigration Sentiment\\
\textbf{source word:} skirt\\
\textbf{source image schema group}: spatial motion\\
\textbf{metaphor implications:} The immigration system is being circumvented through marriage to a brother. Marrying a brother is being used as a method to avoid immigration laws. Marrying a brother to bypass immigration laws is considered a criminal act. The practice of using marriage to circumvent immigration laws should be prohibited.\\

\begin{enumerate}
    \item The \texttt{source\_word} `skirt' is being used metaphorically in the sentence. The writer is pointing to the action of circumventing immigration laws as a crime.
    \item The noun `crime' is best classified as a `thing', so we may set \texttt{correct\_ng} to true. However, `crime' itself is not an `Anti-Immigration Sentiment'. It would be better classified as `Other' because it does not fit cleanly into the target groups provided for the `Immigration' dataset. Therefore we set \texttt{correct\_tg} to false.
    \item The \texttt{source\_word} `skirt', if used literally, would mean to physically pass around something. This fits cleanly into the `spatial\_motion' source schema category. So we mark \texttt{correct\_ss} as true.
    \item The sentences are plausible implications of the metaphorical sentence as a whole, so we mark \texttt{plausible\_implications} as true. The sentences also pertain directly to the metaphorical use of `skirt' in the sentence. Note that the phrase `marrying your brother' here is also a target of the metaphor, as it is being directly considered in terms of the spatial motion. The phrase is imperative to understanding the logic of the metaphor as a whole even though we are focusing here on the `crime' target word. Therefore, the frame implications follow from the logic of the specified metaphorical verb and we may mark \texttt{metaphor\_implications} as true.
\end{enumerate}

\bigskip
\noindent\textbf{\large Example 2}

\smallskip
\textbf{sentence:} So  says there is a perception of Ice being like the KKK that is prohibiting the voting of illegal immigrants.\\
\textbf{target word:} that\\
\textbf{noun group:} organization\\
\textbf{target group:} Klu Klux Klan\\
\textbf{source word:} prohibiting\\
\textbf{source image schema group}: other\\
\textbf{metaphor implications:} Illegal immigrants are being denied voting rights due to restrictive measures. The perception of Ice as resembling the KKK is causing the prohibition of illegal immigrants' voting rights. It is racist and oppressive to equate Ice's actions to the KKK's history of discrimination. The prohibition of illegal immigrants' voting rights should be abolished to address racial injustice.\\

\begin{enumerate}
    \item The \texttt{source\_word} `prohibiting' is being used literally in the context. We acknowledge this, mark \texttt{metaphor\_implications} as `Not Metaphorical', and move on.
    \item The noun `that' in the sentence refers to `the KKK'. It is best classified as a `thing' and fits cleanly into the `Klu Klux Klan' target group. We mark \texttt{correct\_ng} and \texttt{correct\_tg} as true.
    \item The verb `prohibiting' does not cleanly fit into the source schema groups, so `other' is an appropriate annotation. We mark \texttt{correct\_ss} as True.
    \item The sentences in the \texttt{frame\_implications} column do not all agree with the sentence \textit{or even one another}. The sentence itself means that there is a perception that ICE, like the KKK, is disallowing immigrants to vote. It seems implied that this perception is dramatic and false. This does not align with the phrase, ``Illegal immigrants are being denied voting rights due to restrictive measures.'' Therefore, we mark \texttt{plausible\_implications} false and, because `prohibiting' is not being used metaphorically, you should have already marked \texttt{metaphor\_implications} as `Not Metaphorical' (see step 1).
\end{enumerate}

\bigskip
\noindent\textbf{\large Target Word}

\noindent Steps for annotating the \texttt{correct\_ng} (correct noun group) and \texttt{correct\_tg} (correct target group) columns:

\begin{enumerate}
    \item Identify the target word in the context sentence.
    \item Determine if the target word has been correctly defined as person/place/thing/organization and mark the \texttt{correct\_ng} accordingly. If it has, continue to annotate \texttt{correct\_tg}. If not, leave \texttt{correct\_tg} blank.
    \begin{enumerate}
        \item Note that there may be some overlap in the person/thing/organization labels. If the classification is plausible, mark it as correct.
    \end{enumerate}
    \item Navigate to the subtab of this document tab corresponding to the dataset you're working on. This will show the list of possible target groups for each noun group. Determine whether the target group classification is plausible for the target noun; mark \texttt{correct\_tg} accordingly.
\end{enumerate}

\noindent Target groups and definitions for each dataset are included in Tab. \ref{tab:imm-tg}, \ref{tab:gc-tg}, and \ref{tab:a-tg}.

\bigskip
\noindent\textbf{\large Source Word}

\noindent The source image schema group is meant to help characterize the source domain of the metaphor. The goal is to match the metaphorical source verb to the group that best describes the action that the verb would describe if used literally. Note that if you do not agree that the source word is being used metaphorically, you may still annotate this column based on the literal action the verb describes. Details on the groups are included in the `Image Schema Groups' tab.

\begin{enumerate}
    \item Identify the source word in the sentence.
    \item Determine what action the verb would describe if it were used literally.
    \item Does this action plausibly fit into the category determined by the LLM?
    \item The categories and subcategories are not mutually exclusive. In other words, a verb can fit into multiple categories. If the LLM-assigned image schema group is plausible, mark True.
\end{enumerate}

\bigskip
\noindent\textbf{\large Spatial Motion}\\
\textit{physical movement through space}
\begin{itemize}
    \item \textbf{Containment}: being inside or outside a bounded space (e.g., ``trapped'', ``enter'', ``escape'')
    \item \textbf{Path}: moving along a route from one location toward another (e.g., ``head toward'', ``follow'', ``veer'')
    \item \textbf{Source-Path-Goal}: motion with an explicit starting point, route, and destination (e.g., ``depart'', ``arrive'', ``reach'')
    \item \textbf{Blockage}: an obstacle stops or impedes motion (e.g., ``block'', ``stop'', ``hinder'', ``stuck'')
    \item \textbf{Center-Periphery}: position relative to a central core versus the outer edge (e.g., ``marginalize'', ``sideline'', ``encircle'')
    \item \textbf{Cycle}: a repeating loop of motion or events (e.g., ``recur'', ``repeat'', ``loop'')
    \item \textbf{Cyclic Climax}: a repeating pattern that builds toward a peak (e.g., ``escalate'', ``culminate'', ``peak'')
\end{itemize}

\noindent\textbf{\large Force Group}\\
\textit{how forces act on entities}
\begin{itemize}
    \item \textbf{Compulsion}: an external force drives an entity along a path (e.g., ``drive'', ``push'', ``force'')
    \item \textbf{Counterforce}: two opposing forces meet head-on (e.g., ``resist'', ``fight'', ``counter'')
    \item \textbf{Diversion}: a force changes the direction of something in motion (e.g., ``deflect'', ``redirect'', ``divert'')
    \item \textbf{Removal of Restraint}: a restraining force is taken away, releasing motion that was being held back (e.g., ``release'', ``unleash'')
    \item \textbf{Enablement}: a force removes an obstacle or provides the means for motion (e.g., ``enable'', ``allow'', ``empower'')
    \item \textbf{Attraction}: one entity pulls another toward it from a distance (e.g., ``attract'', ``draw in'', ``lure'', ``magnetize'')
    \item \textbf{Link}: a bond or connection constrains how two entities can move relative to each other (e.g., ``bind'', ``tie'', ``connect'')
    \item \textbf{Scale}: the intensity of a force changes (e.g., ``intensify'', ``amplify'')
\end{itemize}

\noindent\textbf{\large Balance Group}\\
\textit{physical equilibrium, drawn from a sense of bodily balance}
\begin{itemize}
    \item \textbf{Axis Balance}: balance around a central vertical axis, as in a body tipping to the side (e.g., ``tip'', ``lean'', ``tilt'')
    \item \textbf{Point Balance}: precarious balance resting on a single pivot point (e.g., ``teeter'', ``poise'', ``balance on'')
    \item \textbf{Twin-Pan Balance}: two sides weighed against each other such as on a scale (e.g., ``weigh'', ``counterbalance'', ``offset'')
    \item \textbf{Equilibrium}: a stable, settled state of rest between competing forces (e.g., ``stabilize'', ``settle'', ``level out'')
\end{itemize}

\noindent\textbf{\large Other}\\
\textit{does not pertain to any of the above groups}

\bigskip
\noindent\textbf{\large Implications}

\begin{enumerate}
    \item First determine whether \textit{all} the sentences listed are plausible implications of the sentence as a whole. If they are, mark \texttt{plausible\_implications} as true. If even one is implausible, mark false and leave \texttt{metaphor\_implications} blank.
    \item If you do not agree that the source word is used metaphorically, leave the \texttt{metaphor\_implications} column blank. Otherwise, determine whether the sentences are specifically related to the metaphor. Sentences will be related to the metaphor if they are guided by the logic of the metaphor. In the \textbf{Example Metaphors} above, the implications must be drawn from the logic of carrying taxes like a heavy burden; it is immoral for less able people to carry it, taxes should be lifted from peoples' shoulders, etc. If they do follow from this logic, mark \texttt{metaphor\_implications} as true. Once again, if even one doesn't plausibly relate to the metaphor, mark false.
\end{enumerate}

}
\end{tcolorbox}

\begin{center}
\small
Codebook 1: Codebook for assessing quality of linguistic metaphor representations.
\label{cb:ling-met-rep}
\end{center}

\twocolumn

%% file: tg_tables.tex
\onecolumn
\begin{center}
\footnotesize
\begin{tabularx}{\textwidth}{|l|X|}
\caption{Target groups for immigration dataset organized by category (Person, Place, Thing, Organization).}
\label{tab:imm-tg} \\
\hline
\textbf{Target Group} & \textbf{Definition} \\
\hline
\endfirsthead

\multicolumn{2}{c}{\tablename\ \thetable{} -- continued from previous page} \\
\hline
\textbf{Target Group} & \textbf{Definition} \\
\hline
\endhead

\hline
\multicolumn{2}{r}{\textit{Continued on next page}} \\
\endfoot

\hline
\endlastfoot

\multicolumn{2}{|l|}{\textbf{Category--Person}} \\
\hline
Immigrants & General population of people who have migrated. \\
Government Officials & Public servants responsible for governance and policy implementation. \\
Political Parties & Organizations that contest elections and influence government policy. \\
Undocumented Immigrants & Individuals in the country without legal authorization. \\
Citizens & Legal residents of a country with rights and responsibilities. \\
Political Figures & Leaders and individuals involved in political processes. \\
Law Enforcement Agencies & Organizations tasked with maintaining public order and safety. \\
Political Leaders & Key decision-makers in political systems. \\
Immigration Reform Advocates & Promote changes in immigration laws and policies. \\
Immigrant Families & Households composed of immigrants and their dependents. \\
Policy Makers & Individuals or groups shaping public policies. \\
People of Color & Demographic group based on racial and ethnic identity. \\
The Establishment & Institutional power structures within political systems. \\
Human Rights Activists & Advocates for the protection and promotion of human rights. \\
Refugees & Individuals who have fled their home countries due to persecution, war, or violence. \\
Political Polarization Supporters & Parties supporting heightened ideological divisions in politics. \\
Court Officials & Judicial personnel including judges and court staff. \\
Voters & Individuals who participate in elections. \\
Border Security Advocates & Supporters of enhanced border control measures. \\
Media Outlets & Organizations disseminating news and information to the public. \\
\hline

\multicolumn{2}{|l|}{\textbf{Category--Place}} \\
\hline
Undocumented Immigrant Communities & Residential areas with significant populations of individuals without legal immigration status. \\
Cities with Poverty, Gangs, and Failed Schools & Urban areas marked by socioeconomic challenges and systemic under-resourced education systems. \\
United States & The nation of focus, characterized by its immigration policies and border dynamics. \\
Mexico & A country in North America with significant immigration ties to the United States. \\
Sanctuary States and Cities & Jurisdictions that restrict cooperation with federal immigration enforcement policies. \\
U.S. Borders & The physical and regulatory boundaries between the United States and neighboring countries. \\
Countries of Origin & Home nations of immigrants, often referenced in discussions about migration patterns and policies. \\
Urban Centers & Major metropolitan areas experiencing significant immigration and demographic shifts. \\
Liberals & Political actors advocating for progressive immigration reform and inclusive policies. \\
Federal Courts & Judicial bodies handling immigration-related legal cases and policy interpretations. \\
Job Magnet Destinations & Regions with robust employment opportunities attracting immigrant labor. \\
Federal Government & The national authority responsible for immigration policy and enforcement. \\
\hline

\multicolumn{2}{|l|}{\textbf{Category--Thing}} \\
\hline
Anti-Immigration Sentiment & Expressed hostility or opposition to immigration, often based on economic, security, or cultural concerns. \\
Legal and Asylum Terminology & Concepts and frameworks around immigration law, asylum eligibility, and legal status distinctions. \\
Border Control and Security & Discussions and strategies related to regulating cross-border movement, enforcement, and physical barriers. \\
Social Benefits and Eligibility & Debates over access to public assistance, healthcare, education, and legal aid for immigrants. \\
Immigration Policy & Policies and frameworks governing immigration, including border security, legal status, and asylum procedures. \\
Economic Impact & Analysis of immigration's effects on labor markets, wages, and public services, both positive and negative. \\
Political Rhetoric and Partisan Debate & Discourse analyzing how political parties and leaders frame immigration issues, including divisive language. \\
Societal Division and Cultural Identity & Discussions about immigration's role in shaping cultural integration, social cohesion, and identity. \\
Humanitarian Concerns & Focus on refugee protection, asylum seekers' rights, and ethical considerations in immigration policy. \\
Criminal Justice and Enforcement & Policies linking immigration status to law enforcement, including detention and deportation practices. \\
Public Opinion and Media Coverage & Analysis of media narratives and public sentiment shaping immigration discourse. \\
Sanctuary City Controversies & Debates over local policies that restrict cooperation with federal immigration enforcement. \\
International Relations & Immigration's role in global diplomacy, bilateral agreements, and transnational migration patterns. \\
Pro-Trump Policies & Support for immigration policies aligned with Donald Trump's administration, including border enforcement and deportation priorities. \\
National Security vs. Humanitarian Concerns & Tensions between immigration enforcement and protecting vulnerable populations. \\
Criticism of Democratic Party & Opposition to Democratic Party platforms on immigration, including border policies and refugee admissions. \\
DACA and Immigration Reform & Discussions around Deferred Action for Childhood Arrivals and broader legislative reforms. \\
Cultural Integration & Processes and challenges of adapting to new cultures, languages, and societal norms. \\
\hline

\multicolumn{2}{|l|}{\textbf{Category--Organization}} \\
\hline
U.S. Government & The federal authority overseeing the governance and administration of the United States. \\
Democratic Party & A major political party in the United States advocating progressive policies and values. \\
Republican Party & A major political party in the United States advocating conservative policies and values. \\
United Nations & An international organization promoting cooperation among nations to address global issues. \\
Religious Organizations & Groups promoting religious beliefs, practices, and community engagement. \\
Major Media Networks & Large-scale news and entertainment organizations shaping public discourse and information dissemination. \\
Ku Klux Klan & A white supremacist organization with a history of promoting racial segregation and violence. \\
U.S. Immigration Enforcement Agencies & Government entities responsible for enforcing immigration laws and border security. \\
Courts & Judicial systems administering justice and interpreting laws within a legal framework. \\
European Union & A political and economic union of 27 European countries with integrated institutions. \\
Tech Companies & Firms engaged in the development and commercialization of advanced technology and digital services. \\
American Civil Liberties Union & A nonpartisan organization advocating for civil rights and constitutional freedoms. \\
Congress & The legislative body of the U.S. federal government responsible for lawmaking. \\
Supreme Court & The highest judicial authority in the United States interpreting the Constitution and federal law. \\
Cabinet & The executive department heads advising the President on various governmental functions. \\
Electoral College & A constitutional mechanism for electing the U.S. President through state-based voting. \\
Mafia Organizations & Criminal syndicates operating in Italy and the U.S., engaged in organized crime and corruption. \\
Cato Institute & A libertarian think tank focused on limited government and free-market policies. \\
Drug Cartels & Organized criminal groups involved in the trafficking of illegal narcotics and other illicit activities. \\
Bolshevik Party & A Marxist political party that led the Russian Revolution and established the Soviet Union. \\
\hline

\end{tabularx}
\end{center}
\clearpage

\begin{center}
\footnotesize
\begin{tabularx}{\textwidth}{|l|X|}
\caption{Target groups for gun control dataset organized by category (Person, Place, Thing, Organization).}
\label{tab:gc-tg} \\
\hline
\textbf{Target Group} & \textbf{Definition} \\
\hline
\endfirsthead

\multicolumn{2}{c}{\tablename\ \thetable{} -- continued from previous page} \\
\hline
\textbf{Target Group} & \textbf{Definition} \\
\hline
\endhead

\hline
\multicolumn{2}{r}{\textit{Continued on next page}} \\
\endfoot

\hline
\endlastfoot

\multicolumn{2}{|l|}{\textbf{Category--Person}} \\
\hline
Political Opponents & Groups or individuals actively opposing specific political agendas or figures. \\
Gun Control Advocates & Individuals and organizations advocating for stricter regulations on firearm ownership and use. \\
Anti-Gun Control Advocates & Individuals and groups opposing gun control measures and promoting firearm accessibility. \\
Legislators Debating Gun Reform & Public officials actively discussing and proposing changes to gun laws. \\
Civil Rights Leaders & Individuals and groups advancing equality and justice for marginalized communities. \\
Political Figures & Elected or appointed leaders shaping public policy and governance. \\
Gun Rights Advocates & Individuals and organizations supporting the right to bear arms and opposing restrictive gun laws. \\
Media Outlets & News organizations and platforms that report on public issues and shape discourse. \\
Families of Gun Violence Victims & Relatives of individuals harmed by firearm-related incidents. \\
Political Parties & Formal organizations representing ideological platforms and electing representatives. \\
Political Strategists & Professionals who plan and execute political campaigns and messaging. \\
Law Enforcement Agencies & Government entities responsible for public safety and crime prevention. \\
Victims of Gun Violence and Their Families & Individuals directly affected by gun violence and their immediate family members. \\
Public Safety Advocates & Groups promoting measures to enhance community security and reduce crime. \\
Public Opinion Survey Respondents & Individuals participating in surveys to gauge societal attitudes on specific issues. \\
Second Amendment Supporters & Individuals and groups endorsing the constitutional right to bear arms. \\
Political Donors and Lobbyists & Individuals or groups influencing policy through financial contributions and advocacy. \\
Activists & Organized groups advocating for specific causes, such as gun safety. \\
Societal Groups Impacted by Gun Violence & Communities and organizations affected by the social and economic consequences of gun violence. \\
Incident Survivors and Their Families & Individuals who survived violent incidents and their families. \\
Military/Paramilitary Groups & Organizations with armed forces or capabilities, including national defense entities. \\
Mental Health Advocates & Individuals and organizations promoting awareness and support for mental health issues. \\
Healthcare Policy Analysts & Experts analyzing and shaping policies related to healthcare access and delivery. \\
Gun Manufacturers & Companies producing firearms and related weaponry. \\
International Comparisons & Entities or analyses comparing U.S. policies with those of other countries. \\
Labor Union Representatives & Organizations representing workers' interests in labor negotiations and policy advocacy. \\
Environmental Advocates & Groups and individuals promoting policies to protect natural resources and combat climate change. \\
Education Reform Supporters & Advocates for systemic changes to improve educational systems and outcomes. \\
\hline

\multicolumn{2}{|l|}{\textbf{Category--Place}} \\
\hline
Cities/Localities & Urban and rural areas managed by local governments for civic services and infrastructure. \\
United States & The sovereign nation comprising 50 states, federal territories, and associated entities. \\
Nation & The collective entity encompassing a country's population, culture, and sovereignty. \\
Other Nations & Countries outside the United States, categorized by geopolitical or economic attributes. \\
Federal Government & The national government at the federal level, including executive, legislative, and judicial branches. \\
Cities with Significant Gun-Related Events & Urban areas with frequent or notable incidents involving firearm violence. \\
Schools & Educational institutions providing primary, secondary, and higher education. \\
Developed Nations & Countries characterized by advanced industrialization, infrastructure, and high standards of living. \\
U.S. States & Administrative divisions within the United States, each with distinct governance and legal systems. \\
State-level Jurisdictions & Legal and administrative divisions within a country, managing local governance and policies. \\
Military/Defense & Institutions and entities responsible for national defense and military operations. \\
Gun Stores & Retail establishments selling firearms and related accessories. \\
Political Offices and Institutions & Roles and frameworks for governing, legislating, and administering public affairs. \\
Countries with Strict Gun Control & Nations enforcing stringent regulations on firearm ownership and use. \\
Mass Shooting Locations & Sites of large-scale firearm attacks, often analyzed for patterns and implications. \\
Congressional Districts & Geographic regions represented by individual members of the U.S. Congress. \\
Gun Rights Organizations & Groups promoting and defending the legal and civil liberties associated with firearm ownership. \\
High-Gun Violence Cities & Urban centers experiencing elevated rates of firearm-related violence. \\
Media/Culture & Media outlets, cultural institutions, and public discourse shaping societal values and narratives. \\
State Governments & Governance structures managing the affairs of individual U.S. states. \\
States with Strict Gun Control Laws & U.S. states implementing rigorous regulations on firearm possession and use. \\
Regions/Zones & Geographic areas defined by cultural, economic, or political characteristics. \\
Religious and Historical Sites & Locations of cultural, spiritual, or historical significance to communities. \\
Public Forums & Open spaces for public debate, discussion, and civic engagement. \\
Political Parties & Organizations advocating for specific ideologies, policies, and electoral strategies. \\
Law Enforcement Agencies & Bodies responsible for maintaining public order, investigating crimes, and enforcing laws. \\
Countries with Permissive Gun Laws & Nations with relaxed legal frameworks regarding firearm ownership and access. \\
\hline

\multicolumn{2}{|l|}{\textbf{Category--Thing}} \\
\hline
Legislative Policy and Implementation & Processes for drafting, passing, and enforcing gun control laws, including challenges to state-level restrictions and federal oversight. \\
Mass Shootings and Tragic Events & Occurrences of large-scale gun violence and their role in catalyzing policy debates, advocacy, and public safety reforms. \\
Political Entities and Advocacy Groups & Organizations and coalitions representing diverse interests, including lawmakers, civil liberties groups, and advocacy networks influencing gun policy. \\
Legal and Constitutional Issues & Legal frameworks, court cases, and constitutional arguments surrounding gun regulation, Second Amendment rights, and federal/state jurisdiction. \\
Political Discourse and Polarization & Debates and discussions within political systems regarding gun control, often influenced by ideological divides and partisan dynamics. \\
Gun Control Policies and Advocacy & Groups and initiatives focused on implementing, promoting, or opposing regulations to restrict firearm access and reduce gun violence. \\
Gun Rights and Civil Liberties Debates & Discussions on constitutional protections for gun ownership, balancing individual rights against public safety concerns. \\
Public Safety Policies and Research & Efforts to enhance community safety through legislative measures, data analysis, and studies on the impact of gun violence on public health and safety. \\
Academic Research and Public Health Impacts & Scholarly studies and health-focused initiatives examining correlations between firearm access, violence, and societal well-being. \\
Public Opinion Trends and Social Media & Shifts in public sentiment toward gun control, driven by digital platforms, grassroots movements, and crisis events. \\
Media Influence and Representation & Role of media in shaping public perception of gun violence, including framing of events, advocacy groups, and political narratives. \\
Cultural Attitudes and Media Framing & Societal beliefs about violence, gun ownership, and trauma, including how cultural narratives and media coverage shape public opinion. \\
Industry and Advocacy Influence & Role of firearm manufacturers, lobbying groups, and advocacy organizations in shaping gun policy and political agendas. \\
Economic Impact of Regulation & Analysis of how gun control policies affect markets, industries, and financial systems, including lobbying efforts and economic incentives. \\
Crisis Response Strategies & Emergency measures and long-term reforms aimed at mitigating gun violence, including trauma support and policy interventions. \\
Historical Legal Precedents & Judicial decisions and legislative milestones shaping the legal landscape of gun rights and regulation over time. \\
International Comparisons and Gun Laws & Global perspectives on firearm regulations, policy effectiveness, and cross-border influences on domestic legislation. \\
\hline

\multicolumn{2}{|l|}{\textbf{Category--Organization}} \\
\hline
National Rifle Association (NRA) & Gun rights advocacy organization promoting Second Amendment protections and firearm ownership. \\
Government Law Enforcement Agencies & Federal and state entities responsible for enforcing gun laws and investigating violations. \\
Bureau of Alcohol, Tobacco, Firearms and Explosives (ATF) & Agency regulating firearms and enforcing federal gun laws. \\
Republican Party & Political party emphasizing Second Amendment rights and limited government regulation. \\
Democratic Party & Political party advocating progressive policies, including gun control measures. \\
Supreme Court & Highest judicial body interpreting gun rights and regulations under the U.S. Constitution. \\
Federal Courts & Judicial branch overseeing legal cases involving gun-related laws and constitutional challenges. \\
Public Health Research Institutions & Academic and research entities analyzing gun violence trends and health impacts. \\
Gun Rights Advocacy Organizations & Collective term for groups defending Second Amendment rights and firearm freedoms. \\
Centers for Disease Control and Prevention (CDC) & Public health agency conducting research on firearm-related injuries and mortality. \\
Local Law Enforcement Agencies & Community-based police departments and sheriff's offices enforcing gun regulations. \\
Gun Violence Research Organizations & Institutions studying the causes and impacts of firearm violence for policy development. \\
Gun Control Advocacy Groups & Organizations promoting policies to regulate firearms and reduce gun violence. \\
Nonprofit Organizations & Charitable entities working on gun safety, education, and community programs. \\
Federal Bureau of Investigation (FBI) & Agency investigating federal firearm crimes and enforcing gun control laws. \\
Gun Violence Prevention Coalition & Collective of groups working to implement gun safety measures and violence prevention strategies. \\
Gun Control Advocacy Organizations & Collective term for groups promoting firearm regulation and safety measures. \\
Brady Campaign to Prevent Gun Violence & Advocacy group focusing on universal background checks and firearm regulation. \\
Policy Advocacy Groups & Organizations influencing legislation through research, lobbying, and public campaigns. \\
Moms Demand Action & Grassroots movement advocating for gun safety and legislative reforms. \\
American Civil Liberties Union (ACLU) & Civil liberties advocacy group defending constitutional rights, including Second Amendment protections. \\
Campaign to Stop Gun Violence & Grassroots movement advocating for stricter gun control laws. \\
Firearms Safety Organizations & Groups providing education and resources on safe firearm handling and storage. \\
March for Our Lives & Youth-led advocacy group pushing for gun violence prevention policies. \\
Second Amendment Foundation & Organization focused on preserving constitutional gun rights and legal advocacy. \\
\hline

\end{tabularx}
\end{center}
\clearpage

\begin{center}
\footnotesize
\begin{tabularx}{\textwidth}{|l|X|}
\caption{Target groups for abortion dataset organized by category (Person, Place, Thing, Organization).}
\label{tab:a-tg} \\
\hline
\textbf{Target Group} & \textbf{Definition} \\
\hline
\endfirsthead

\multicolumn{2}{c}{\tablename\ \thetable{} -- continued from previous page} \\
\hline
\textbf{Target Group} & \textbf{Definition} \\
\hline
\endhead

\hline
\multicolumn{2}{r}{\textit{Continued on next page}} \\
\endfoot

\hline
\endlastfoot

\multicolumn{2}{|l|}{\textbf{Category--Person}} \\
\hline
Republicans & Political party members who support conservative policies and values. \\
Women & Adult females, often analyzed for their unique experiences and roles in societal issues. \\
Political commentators & Media personalities and analysts who interpret political events and policies. \\
Pro-life advocates & Individuals who oppose abortion and advocate for the protection of fetal life. \\
Religious conservatives & Individuals who emphasize traditional religious values, including the role of God in public life. \\
Pro-choice advocates & Individuals who support the right to access abortion services and reproductive healthcare. \\
Religious leaders & Spiritual figures such as pastors, imams, rabbis, and priests who lead religious communities. \\
Public figures & Highly visible individuals in politics, entertainment, or other fields who influence public opinion. \\
Public opinion on abortion rights & General attitudes and beliefs about the ethical and legal status of abortion. \\
Men & Adult males, often analyzed for their distinct societal roles and perspectives. \\
Policy makers & Government officials and legislators responsible for creating and implementing policies. \\
Judicial figures & Legal professionals including judges, attorneys, and court officials who interpret laws. \\
Democrats & Political party members who support progressive policies and social reforms. \\
Political ideologies & Belief systems such as communism, socialism, and libertarianism that shape policy perspectives. \\
Activists in social movements & Participants in organized efforts to achieve specific social, political, or cultural change. \\
Media outlets & News organizations and platforms that shape public discourse and information dissemination. \\
Abortion providers & Healthcare professionals who offer abortion services and reproductive care. \\
Religious organizations & Institutions and groups that promote and practice specific religious beliefs and practices. \\
Medical professionals & Doctors, nurses, and other healthcare workers involved in patient care and treatment. \\
Religious vs. secular perspectives & Divergent viewpoints on the role of religion in governance and public policy. \\
Social justice activists & Individuals and groups working to address systemic issues like racism, inequality, and discrimination. \\
Advocacy groups & Organizations that promote specific causes, such as human rights or environmental protection. \\
Reproductive rights organizations & Groups focused on expanding access to abortion, contraception, and sexual health services. \\
Healthcare access and equity & Issues related to the availability and fairness of healthcare services. \\
Legal and policy debates & Discussions around the interpretation and implementation of laws and regulations. \\
Political strategies and elections & Processes and tactics used in political campaigns and electoral systems. \\
\hline

\multicolumn{2}{|l|}{\textbf{Category--Place}} \\
\hline
Religious Institutions & Organizations or entities focused on religious practices, teachings, and community service. \\
Healthcare Facilities & Establishments providing medical services and care, including hospitals and clinics. \\
Nations & Independent countries or sovereign states recognized globally. \\
Global Context & International perspectives or data comparisons that provide a worldwide framework for understanding the issue. \\
Restrictive Abortion Laws States & States with legislation that imposes significant restrictions on access to abortion services. \\
Foreign Countries & Non-U.S. nations involved in international relations or data comparisons. \\
Midwest States & States located in the central region of the United States, often characterized by agricultural economies and industrial centers. \\
National Entities & Governments or sovereign states that are recognized internationally. \\
Pro-Life Cities & Urban areas with policies or cultural emphasis on opposing abortion. \\
U.S. States & Administrative divisions within the United States with autonomous governing structures. \\
Urban Areas & Populous, densely populated regions characterized by high population density and infrastructure development. \\
Southern States & States in the southern part of the United States, often characterized by distinct cultural, economic, and political profiles. \\
Educational Institutions & Schools, colleges, and universities providing academic and professional training. \\
State Governments & Administrative bodies responsible for governing individual U.S. states. \\
Liberal Abortion Laws States & States with legislation that ensures broad access to abortion services. \\
Rural Areas & Sparsely populated regions with low population density and limited infrastructure development. \\
Protective Zones & Areas designated for safeguarding critical infrastructure, populations, or resources. \\
Progressive Cities & Urban centers known for supporting social, environmental, and political progressive policies. \\
U.S. Political Offices & Positions within the U.S. government at the federal, state, or local level. \\
Federal Legislative Bodies & Institutions at the national level responsible for creating and passing laws. \\
New York City & A major urban center in the United States known for its economic, cultural, and political influence. \\
Political Swing States & States where electoral outcomes are uncertain and can shift between political parties. \\
Electoral Districts & Geographic divisions used to allocate political representation in elections. \\
\hline

\multicolumn{2}{|l|}{\textbf{Category--Thing}} \\
\hline
Religious/Moral Perspectives & Beliefs and arguments from religious or moral frameworks regarding abortion and fetal life. \\
Political Parties & Divergent policy priorities and advocacy strategies between major U.S. political parties. \\
Political Debates & Discussions and conflicts between political ideologies on abortion, healthcare, and related policies. \\
Abortion Access and Legal Rights & Groups focused on expanding or restricting access to abortion services and legal protections for reproductive rights. \\
Reproductive Rights Advocacy & Organizations and movements advocating for women's autonomy in reproductive healthcare decisions. \\
Political Campaigns and Legislation & Strategies and legislative efforts to influence abortion laws and healthcare funding. \\
Media and Public Opinion Influence & Role of media in shaping narratives and public sentiment on abortion and healthcare issues. \\
Legislative Proposals (Heartbeat Act, 20-Week Bans) & Specific bills targeting restrictions on abortion at certain gestational stages. \\
Demographic Data and Trends & Analysis of population statistics influencing policy debates and healthcare resource allocation. \\
Fetal Rights Advocacy & Promotion of legal protections for fetal life, often opposing abortion access. \\
Legal and Judicial Precedents & Judicial rulings and legal frameworks shaping abortion regulations and reproductive rights. \\
Social Justice Issues & Overlaps between reproductive rights, racial equity, and other systemic inequalities. \\
Political Movements & Broad-based campaigns uniting diverse groups around abortion rights or restrictions. \\
Civil Rights Movements & Historical and contemporary efforts to protect rights related to race, gender, and reproductive freedom. \\
Constitutional Law and Judicial Review & Interpretations of the Constitution's role in regulating abortion and healthcare policies. \\
Women's Rights Advocacy & Movements promoting gender equality, including reproductive healthcare and labor rights. \\
Healthcare Access and Equity & Issues surrounding disparities in healthcare access and the impact of policy decisions on marginalized groups. \\
Legal Challenges and Court Rulings & Litigation and judicial decisions testing the constitutionality of abortion restrictions. \\
Healthcare Organizations (Planned Parenthood, ACLU) & Institutions providing reproductive healthcare services and advocating for policy reform. \\
Public Health and Medical Ethics & Focus on ethical considerations and health implications of reproductive healthcare policies. \\
Feminism and Gender Equality & Advocacy for women's rights, including reproductive autonomy and workplace equity. \\
Ethical and Scientific Debates & Discussions on the scientific and ethical status of fetal life and reproductive healthcare. \\
Healthcare System Reform & Proposals to restructure healthcare delivery and funding mechanisms for reproductive services. \\
Public Opinion Polarization & Divergence in public attitudes toward abortion, influenced by political and cultural factors. \\
Economic and Social Implications & Analysis of how abortion policies affect healthcare costs, labor markets, and social structures. \\
Health Disparities & Inequalities in healthcare access and outcomes based on race, class, or geography. \\
International Comparative Perspectives & Comparative analysis of abortion laws and policies across countries. \\
Religious Freedom and Civil Liberties & Conflicts between religious beliefs and secular legal protections for reproductive rights. \\
\hline

\multicolumn{2}{|l|}{\textbf{Category--Organization}} \\
\hline
Political Parties & Political entities representing opposing ideologies in U.S. governance. \\
Media Outlets & News organizations influencing public discourse, including major publications like The New York Times and Breitbart. \\
Government Agencies & Federal bodies responsible for administering policies and regulations. \\
Congress & The legislative branch of the U.S. federal government. \\
Anti-Abortion Advocacy Groups & Organizations opposing abortion access, such as Concerned Women for America and the Center for Medical Progress. \\
Political Lobbying Groups & Organizations advocating for specific policy agendas through influence on legislation. \\
Religious Organizations & Faith-based groups influencing public policy, including the Catholic Church, Vatican, Ecumenical Coalition on Women and Society. \\
Reproductive Health Organizations & Institutions providing reproductive healthcare services and education, including Planned Parenthood, Guttmacher Institute, and National Family Planning and Reproductive Health Association. \\
Legal/Legislative Process & Procedural mechanisms for enacting laws and adjudicating legal cases. \\
Abortion Rights Advocacy & Organizations and individuals promoting access to legal abortion services. \\
Non-Profit Advocacy Groups & Civil society organizations addressing social issues, such as the American Civil Liberties Union (ACLU) and the National Organization for Women (NOW). \\
Legal Defense Organizations & Entities providing legal support for advocacy and litigation, including the ACLU. \\
Education Policy & Initiatives addressing curriculum standards and school reforms, including Common Core-related efforts. \\
Abortion Policy Advocacy & Groups focused on shaping legislative and regulatory frameworks around abortion access. \\
\hline

\end{tabularx}
\end{center}
\clearpage

\begin{center}
\footnotesize
\begin{tabularx}{\textwidth}{|l|X|}
\caption{Target groups for podcast dataset organized by category (Person, Place, Thing, Organization).}
\label{tab:p-tg} \\
\hline
\textbf{Target Group} & \textbf{Definition} \\
\hline
\endfirsthead

\multicolumn{2}{c}{\tablename\ \thetable{} -- continued from previous page} \\
\hline
\textbf{Target Group} & \textbf{Definition} \\
\hline
\endhead

\hline
\multicolumn{2}{r}{\textit{Continued on next page}} \\
\endfoot

\hline
\endlastfoot

\multicolumn{2}{|l|}{\textbf{Category--Person}} \\
\hline
Voters & Eligible citizens who can participate in elections. \\
Candidates & Individuals seeking elected office in government positions. \\
Minority Communities & Ethnic, racial, or cultural groups with smaller populations, often facing systemic challenges. \\
Politicians & Individuals holding elected or appointed public office, including legislators, executives, and officials. \\
Palestinians & Individuals from Palestine, a region with complex political and historical contexts. \\
MAGA Republicans & A conservative political group within the Republican Party advocating for Trump-aligned policies. \\
Law Enforcement & Officers and agents responsible for maintaining public safety and enforcing laws. \\
Workers & Individuals engaged in employment to earn a living across various industries. \\
Media & News organizations, journalists, and commentators providing public information and analysis. \\
Democrats & A liberal political party in the U.S. \\
Judges & Judicial officers responsible for interpreting and applying laws. \\
Online Commentators & Individuals engaging in public discussion and commentary on digital platforms. \\
Socialists & Advocates for socialist principles, including economic equality and government control of production. \\
Defense Officials & Government personnel overseeing national defense and military operations. \\
Students & Individuals enrolled in educational institutions pursuing academic or vocational training. \\
Republicans & A conservative political party in the U.S. \\
Activists & Individuals or groups advocating for social, political, or environmental change. \\
Military Personnel & Members of the armed forces engaged in national defense and security. \\
Lawmakers & Legislators and officials tasked with drafting, debating, and enacting laws. \\
January 6th Insurrectionists & Individuals involved in the 2021 Capitol attack, a violent breach of the U.S. Capitol. \\
Immigrants & Individuals who have relocated to a new country for residence, including legal and undocumented residents. \\
Disenfranchised Communities & Groups excluded from political processes or facing barriers to civic participation. \\
Corporate Executives & Senior leaders managing business operations and strategic direction. \\
Religious Leaders & Guides of religious institutions or communities, leading spiritual and ethical practices. \\
Protesters & Participants in public demonstrations to express opposition or support for a cause. \\
Environmentalists & Advocates for the protection, conservation, and sustainable use of natural environments. \\
Entrepreneurs & Business owners or innovators launching new ventures to provide goods or services. \\
Healthcare Professionals & Medical practitioners and allied health workers providing patient care and treatment. \\
European Allies & European countries or officials with strong political, economic, or military ties to the U.S. \\
Shareholders & Individuals or entities owning shares in a company, representing partial ownership. \\
Border Patrol Agents & Federal officers enforcing immigration and border security laws. \\
Academics & Scholars and researchers engaged in teaching and research at educational institutions. \\
Teachers & Educators who instruct students in schools or other educational settings. \\
\hline

\multicolumn{2}{|l|}{\textbf{Category--Place}} \\
\hline
Urban Areas & Densely populated regions with significant economic and cultural hubs, encompassing cities and their surrounding metropolitan areas. \\
Washington, D.C. & The federal district and capital of the United States, functioning as a distinct political and administrative entity. \\
Midwestern States & U.S. states in the central region, often associated with agriculture, manufacturing, and mid-sized urban areas. \\
Red States & States with predominantly conservative political leanings and policies. \\
Latin American Countries & Nations in Latin America, encompassing diverse political, economic, and cultural contexts. \\
Western States & U.S. states in the western region, including mountainous and arid landscapes with diverse economies and political landscapes. \\
Southern States & U.S. states in the southeastern and south-central regions, historically associated with specific cultural, economic, and political characteristics. \\
Blue States & States with predominantly liberal political leanings and policies. \\
Mountain States & U.S. states in the mountainous western region, characterized by rugged terrain and specific resource-based economies. \\
Key Congressional Districts & Elected districts with significant influence on legislative outcomes or contested in national elections. \\
\hline

\multicolumn{2}{|l|}{\textbf{Category--Thing}} \\
\hline
Political Violence & Acts of violence linked to political ideologies or power struggles. \\
History & Recorded events and their analysis to understand past influences on the present. \\
Media and Public Figures & Influential individuals or entities shaping public discourse through media. \\
Political Campaigns & Strategies and activities aimed at promoting political candidates or initiatives. \\
Political Strategy & Plans and tactics used to achieve political objectives or influence decision-making. \\
Legal Matters & Issues involving legal systems, regulations, and judicial processes. \\
Research & Systematic investigation into a subject to discover new knowledge or solutions. \\
Economy & The system through which goods, services, and resources are produced, distributed, and consumed. \\
Social and Political Issues & Topics intersecting social challenges with governance and policy. \\
Legal and Criminal Proceedings & Judicial processes to address alleged crimes or legal violations. \\
Education & Processes and institutions aimed at imparting knowledge and skills. \\
Privacy & The right to keep personal information confidential and control its disclosure. \\
Economic Issues & Topics related to economic policies, growth, and resource distribution. \\
Public Policy Debates & Discussions on the creation, implementation, or evaluation of public policies. \\
Fiscal Policy & Government strategies to manage taxation, spending, and public debt. \\
Data & Information collected, stored, and analyzed to inform decisions or research. \\
Government & The system or body exercising authority over a state or territory. \\
Community & Groups of people sharing common interests, locations, or social ties. \\
Media and Public Perception & The role of media in shaping public opinion and societal views. \\
Healthcare & Services and systems focused on medical care, prevention, and public health. \\
Media Events & Significant occurrences covered by media, influencing public discourse. \\
Misinformation & The spread of false or misleading information through media channels. \\
International Relations & Interactions between nations, including diplomacy and treaties. \\
Inaction & The failure to take necessary steps or respond to a situation effectively. \\
Critical Issues & Urgent problems requiring immediate attention or reform. \\
Security & Measures to protect individuals, assets, or systems from threats or harm. \\
Law Enforcement & Issues surrounding criminal activities, policing, and justice administration. \\
Personal Perspectives & Individual viewpoints shaped by personal experiences and beliefs. \\
Public & The general population or citizens of a society or region. \\
Long-term Trends & Patterns of development or change observed over extended periods. \\
Political Movements and Parties & Organized groups advocating for specific political ideologies or reforms. \\
Corporate Governance & Practices ensuring accountability, transparency, and ethical behavior in business leadership. \\
Policy & Rules or guidelines established to govern actions or decisions within an organization or society. \\
Testimony & Statements or evidence provided by individuals in legal or investigative contexts. \\
Social Inequality & Differences in wealth, opportunity, or power among social groups. \\
Political Cycles & Recurring patterns in political behavior, power dynamics, or electoral outcomes. \\
Economic Disparities & Differences in economic conditions between regions or demographic groups. \\
Government Investigations & Processes by which government bodies examine misconduct or policy violations. \\
Technology & Tools, systems, and methods developed to enhance human capabilities and efficiency. \\
Care & Systems and practices supporting health, well-being, and daily living assistance. \\
Environment & The natural world and its ecosystems, including air, water, and land. \\
Digital Privacy Concerns & Issues related to the protection of personal data in digital environments. \\
Demographics & Statistical data on population characteristics such as age, gender, or ethnicity. \\
Society & The collective way of life, customs, and institutions of a community or group. \\
Ethics & Principles governing moral behavior and decision-making in professional or personal contexts. \\
Religious Extremism & Ideologies advocating extreme views or actions based on religious beliefs. \\
\hline

\multicolumn{2}{|l|}{\textbf{Category--Organization}} \\
\hline
Disney & Multinational entertainment conglomerate producing media, theme parks, and consumer products. \\
Democratic National Committee & National governing body of the Democratic Party. \\
Palestine Liberation Organization & Political organization advocating for Palestinian statehood and self-determination. \\
Democratic Party & Political party representing progressive ideologies and policies in the U.S. \\
Republican Party & Political party representing conservative ideologies and policies in the U.S. \\
State Department & Federal agency managing foreign affairs and international relations. \\
CBS & Major television network providing news, entertainment, and media content. \\
Congress & Bicameral legislative body enacting federal laws and approving budgets. \\
Office of the President of the United States & Executive branch leadership role overseeing national policy and governance. \\
Supreme Court & Highest judicial body interpreting U.S. Constitution and federal law. \\
Federal Reserve System & Central banking system managing monetary policy and financial stability. \\
Department of Defense & Federal agency overseeing national defense and military operations. \\
Department of Justice & Federal executive department overseeing legal matters, prosecutions, and justice administration. \\
Culinary Union & Labor organization representing restaurant and hospitality industry workers. \\
Mossad & Israeli intelligence agency responsible for national security operations. \\
Federal Bureau of Investigation (FBI) & Federal law enforcement agency focused on counterterrorism, cybercrime, and criminal investigations. \\
Department of Homeland Security & Federal agency managing border security, counterterrorism, and disaster response. \\
Central Intelligence Agency & Federal agency conducting intelligence gathering and counterintelligence operations. \\
state of Texas & U.S. state with distinct political, cultural, and economic characteristics. \\
White House Correspondents Association & Professional organization representing press corps covering the U.S. presidency. \\
U.S. Immigration and Customs Enforcement (ICE) & Federal agency responsible for immigration enforcement and border security. \\
\hline

\end{tabularx}
\end{center}